\documentclass[a4paper]{report}
\usepackage[utf8]{inputenc}
\usepackage[T1]{fontenc}
\usepackage{RJournal}
\usepackage{amsmath,amssymb,array}
\usepackage{booktabs}

\usepackage{makecell}
\usepackage{multirow}
\def\rot#1{\rotatebox{90}{#1}}
\usepackage{amssymb}% http://ctan.org/pkg/amssymb
\usepackage{pifont}% http://ctan.org/pkg/pifont
\newcommand{\cmark}{\ding{51}}%
\usepackage{tablefootnote}
\usepackage{pdflscape}
\usepackage{makecell}
\usepackage{lscape}

\PassOptionsToPackage{hyphens}{url}\usepackage{hyperref}

\begin{document}

%% do not edit, for illustration only
\sectionhead{Contributed research article}
\volume{XX}
\volnumber{YY}
\year{20ZZ}
\month{AAAA}

%% replace RJtemplate with your article
\begin{article}
  % !TeX root = RJwrapper.tex

\title{Landscape of R packages for eXplainable Artificial Intelligence}
\author{by Szymon Maksymiuk, Alicja Gosiewska, Przemysław Biecek}

\maketitle

\abstract{
The growing availability of data and computing power is fueling the development of predictive models. In order to ensure the safe and effective functioning of such models, we need methods for exploration, debugging, and validation. New methods and tools for this purpose are being developed within the eXplainable Artificial Intelligence (XAI) subdomain of machine learning.
In this work, (1)~we present the taxonomy of methods for model explanations, (2) we identify and compare 27~packages available in R to perform XAI analysis, (3) we present an example of an application of particular packages, (4) we acknowledge trends in recent developments. The article is primarily devoted to the tools available in R, but since it is easy to integrate the Python code, we will also show examples for the most popular libraries from Python.
}

%\section{Introduction}

\section{Importance of  eXplainable Artificial Intelligence}
\label{sec:importance}

The growing demand for fast and automated development of predictive models has contributed to the popularity of machine learning frameworks such as \CRANpkg{caret} \citep{caret}, \CRANpkg{mlr} \citep{mlr}, \CRANpkg{mlr3} \citep{mlr3}, \CRANpkg{tidymodels} \citep{tidymodels}, \pkg{h2o} \citep{h2o}, \pkg{scikit-learn} \citep{sklearn}, \pkg{keras} \citep{keras}, \pkg{pytorch} \citep{pytorch}, and many others.
These tools allow us to quickly test models of very different structures and choose the best one based on a selected performance measure.  
However, it soon became apparent that such a process leads to treating models like black-boxes. This, in turn, makes it difficult to detect certain problems early enough.
Insufficiently tested models quickly lose their effectiveness, lead to unfair decisions, discriminate, are deferred by users, and do not provide the option to appeal \citep{responsibleml}.
To overcome these problems methods and tools for analysis of black-box predictive models are essential. In this work, we will present tools that can be used for this purpose.

There are various situations where we need tools for in-depth model analysis. For example:

\begin{itemize}
    \item  The model makes incorrect decisions on certain observations. We want to understand what the cause is of these incorrect decisions, in hopes to improve the model. In some sense, we want to debug the model by looking for the source of its ineffectiveness.
    \item Model predictions are used by inquisitive parties. In order to build their trust and confidence, we need to present additional arguments or reasoning behind particular predictions.
    \item Sometimes we expect that the model will automatically discover some relationships hidden in the data. By analyzing the model, we want to extract and understand these relationships in order to increase our domain knowledge.
    \item It is becoming increasingly common to expect not only decisions but also reasons, arguments, and explanations for a decision to be made in an automated way. Sometimes such explanations are required by law, see, for example, GDPR regulations \citep{Goodman_Flaxman_2017}.
    \item Often the key factor is the question of responsibility. If model developers responsibly recommend the use of a model, they often need to understand the way the model works. Thus, we cannot rely on the black-box. We need a deeper understanding of the model.
\end{itemize}

In this work, we present tools that can be used in eXplainable Artificial Intelligence (XAI) modeling, which can help to explore predictive models. In recent years, plenty of interesting software has been developed. We hope that presenting XAI tools in this paper will make them easier to apply and will, therefore, lead to building better and safer predictive models.
The main contributions of this paper are as follows.
\begin{enumerate}
    \item We introduced two taxonomies of methods for model comparisons. The first one concerns a~model as a subject of analysis, and the second one concerns the domain of the explanation.  
    \item We conducted a comprehensive review and identified 27 popular R packages that implement XAI methods.
    \item We compared the capabilities of recognized R packages, taking into account the variety of implemented methods, interoperability, and time of operation. 
    \item We prepared knitr reports with illustrative examples for each of the considered packages.
\end{enumerate}

This paper is focused on the tools available to R users rather than on an overview of mathematical details for particular methods. Those interested in a more detailed analysis of the methods may want to familiarize themselves with a work of \citet{doi:10.1177/1473871620904671} with a meta-analysis of 18 survey papers that refer to the explainability of machine learning models. Some of them are related to model visualization, such as  \citet{LIU201748}, predictive visual analytics \citep{lu2017recent,doi:10.1111/cgf.13210}, interaction with models \citep{Amershi_Cakmak_Knox_Kulesza_2014, dudley2018review}, deep learning \citep{choo2018visual,GARCIA201830,DBLP:journals/corr/abs-1801-06889}, or dimensionality reduction \citep{sacha2016visual}. Algorithmics details for XAI methods can also be found in books \textit{Interpretable Machine Learning} \citep{molnar2019} or \textit{Explanatory Model Analysis}  \citep{ema}. \citet{doi:10.1177/1473871620904671} emphasized the importance of regularly keeping up with new surveys, both due to the covering of more articles as well as the different perspectives. 
To the best of our knowledge, none of these surveys aims at a comprehensive analysis of software tools for eXplainable Artificial Intelligence. In this paper, we conduct an extensive review of the R packages.

It should be noted that work in this area is being carried out in various camps of modelers. The contributions of the statisticians are intertwined with those made by practitioners in machine learning or deep learning. This sometimes leads to redundancy in the nomenclature. The name \textit{eXplainable Artificial Intelligence} was popularized by the DARPA program \citep{darpa}, emphasizing the question of how much the algorithm can explain the reasons for a recommended decision. The name \textit{Interpretable Machine Learning} was popularized by a book with the same title \citep{molnar2019}, which emphasizes a model's interpretability. 
The name \textit{Explanatory Model Analysis} \citep{ema} refers to the main objective, which is to explore model behavior. These threads are also known as \textit{Veridical Data Science} \citep{Yu3920} or \textit{Responsible Machine Learning} \citep{responsibleml}. In this paper, we will use the term XAI, but note that in most cases these names can be used interchangeably.

\section{Taxonomy of XAI methods}
\label{sec:taxonomy}

\begin{figure}[htb]
  \centering
  \includegraphics[width=\textwidth]{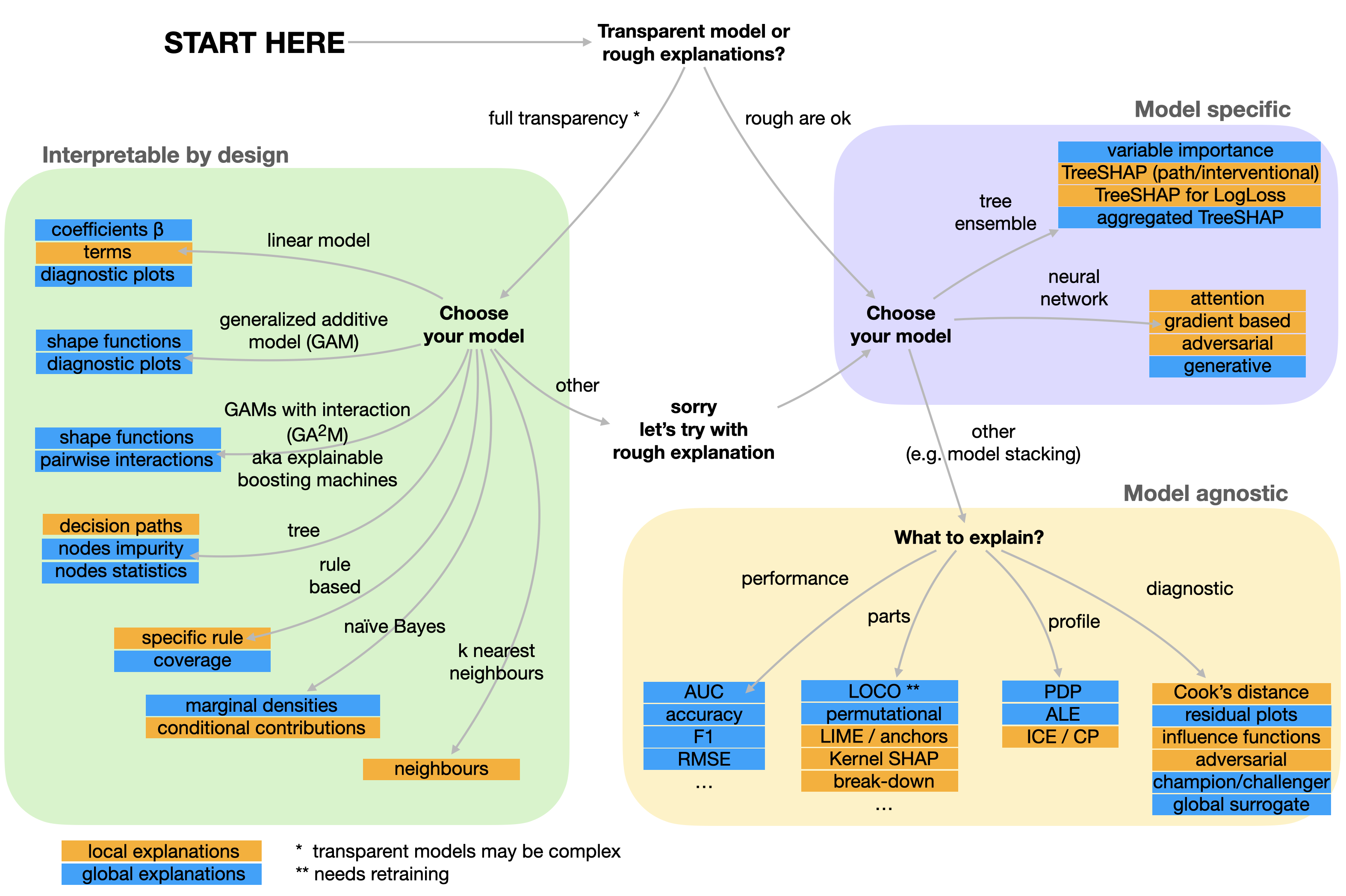}
  \caption{Model oriented taxonomy for XAI methods.}
  \label{fig:map}
\end{figure}

The literature presents several taxonomies of XAI methods categorization \citep{XAItaxonomies, biran2017explanation, molnar2019, ema}. Figure~\ref{fig:map} summarizes the most frequent grouping of these methods.

The introduced taxonomy puts the model structure in the center. From this point of view, we are dealing with three groups of methods.

\begin{itemize}
    \item Methods for models with interpretable structure (\textbf{interpretable by design}), such as linear models, decision trees \citep{partyctree}, decision rules \citep{arules}, k-nearest neighbors \citep{knn,kknn}, and others. The architecture of such predictive models allows us to directly explain a certain aspect of the model. Such explanations are accurate, i.e. they are based on model coefficients and describe the model in a complete way. Of course, such models can be difficult to understand. A tree can contain hundreds of decision leaves, making it challenging to grasp the whole. But, the model structure is, in general, directly interpretable.
    \item \textbf{Model-specific methods}. There are situations in which the model structure is so complex that it cannot be interpreted by looking at individual parameters. However, there are methods for knowledge extraction designed for specific classes of models. For tree ensembles, it is possible to analyze the tree structures in order to extract from them some relationships between variables. For neural networks, it is possible to trace gradients. This group of methods assumes full access to the model structure. We expect that the explanation of the operation is a rough approximation of a more complex process.
    \item \textbf{Model-agnostic methods}. The most general class of methods are those that facilitate analysis of the model without any knowledge of the internal structure. The analysis is usually carried out on the basis of a number of model evaluations on properly prepared perturbed input data. Such an analysis can be carried out on any predictive model. 
\end{itemize}

In the last chapter of this paper, we will present examples of methods for each of these groups, but the main emphasis is on model-agnostic methods. The reason for this is that we often compare models with different structures when looking for a final model, and, sometimes, we even have to compare models created in different frameworks. The main advantage of that approach is flexibility. It allows users of such an algorithm to compare explanations of different machine learning models, for instance, the forest type model with a boosting tree or even neural network, with common metrics between them. The disadvantage is that explanations acquired with that approach may be less accurate. They only approximate the behavior of the predictive model ignoring information that comes from the structure.

Another possible taxonomy refers to the task to which the explanation is used. The most common are two situations when the explanation concerns one observation or the whole dataset. If the explanation concerns a single observation, it is called \textbf{local} or individual. If an explanation refers to the whole dataset, it is called \textbf{global} or dataset-level. In applications, we sometimes fall into a~gray zone where we are interested in the behavior of the model for a group of observations. Usually, both local and global methods can be used in such cases.

Table~\ref{tab:hierarchy} introduces the second taxonomy of methods according to the purpose of model explanation, whether the goal is to understand the importance of the variables, profile the variables, or to analyze the performance of a model. Later on in the article, we will present examples of comparisons of models based on this taxonomy.

\begin{table}[htb]
\centering
\begin{tabular}{|l|l|}
\multicolumn{1}{c}{\textit{Global}} & \multicolumn{1}{c}{\textit{Local}} \\ 
\hline
\begin{tabular}[c]{p{6cm}p{6cm}}\textbf{Model Parts}\\\\ $\bullet$ Permutational variable importance \\ $\bullet$ Leave-One-Covariate-Out (LOCO)\\$\bullet$ Surrogate models \\ $\bullet$ Aggreagated SHapley Additive ex-\\ \ \  \ Planations \\ \\ \end{tabular} & \begin{tabular}[c]{p{6cm}p{6cm}}\textbf{Predict Parts}\\\\ $\bullet$ BreakDown (BD) \\ $\bullet$ SHapley Additive exPlanations \\ \ \ \  (SHAP) \\ $\bullet$ Local Interpretable Model-agnostic \\ \ \ \  Explanations (LIME) \\ \\ \end{tabular} \\ 
\hline
\begin{tabular}[c]{p{6cm}p{6cm}}\textbf{Model Profile}\\ \\ $\bullet$ Partial Dependence Profiles (PDP) \\ $\bullet$ Accumulated Local Effects (ALE)\end{tabular} & \begin{tabular}[c]{p{6cm}p{6cm}}\textbf{Predict Profile}\\\\$\bullet$ Ceteris Paribus (CP) / Individal\\ \ \ \ Conditional Expectations (ICE)\end{tabular} \\ 
\hline
\begin{tabular}[c]{p{6cm}p{6cm}}\textbf{Model Diagnostics}\\\\$\bullet$ Residual plots \\ $\bullet$ Variable vs. prediction plots \\ $\bullet$ Demographic parity\end{tabular} & \begin{tabular}[c]{p{6cm}p{6cm}}\textbf{Predict Diagnostics}\\\\$\bullet$ Local residual density plot \\ \\ \\ \end{tabular} \\
\hline
\end{tabular}

\caption{XAI taxonomy for model-agnostic and model-specific methods along with examples. Names in \textit{italics} at the top denote the explanation domain. \textbf{Bolded} names refer to the explanation task, while itemized entries are examples of explanation methods that comply with the given task.}
\label{tab:hierarchy}
\end{table}

\clearpage

Below we discuss the objectives for each task and give examples of methods.

\begin{itemize}
    \item \textbf{Model Parts} focuses on the importance of variables or groups of variables. It considers a~global level approach. The Model Parts task can be realized by computing the impact of a~single variable or group of variables have on the model performance using various methods like permutation variable importance,  Leave One Covariate Out (LOCO) \citep{loco}, or surrogate tree models. It seems the most popular is the permutation variable importance \citep{fi} which assesses the importance of a variable by the change in the model performance after masking the effect of one or a group of variables. The effect of masking is achieved by permutations or resampling values for selected variables.
    
    \item \textbf{Model Profile} aims at presenting how a single variable or a group of variables affects model response. It shows the profile of model prediction as a function of dependent variables. Model Profile represents the global domain of explanations, where the whole dataset is taken into consideration. The most popular examples of such methods are Partial Dependence Profiles (PDP) and Accumulated Local Effects (ALE).
    PDP \citep{Friedman00greedyfunction} is a method of profiling the global behavior of models in the context of one or a pair of variables. Its purpose is to show how the expected prediction of the model change based on changes to dependent variables. In other words, it estimates model response as a function of a specific variable. 
    ALE  \citep{2016arXiv161208468A} is to some extent, similar to PDP, but it takes into account the conditional distribution of a~variable, instead of a marginal one. Thus, it is more suited in cases where variables are correlated. 
    \item \textbf{Model Diagnostics} focuses on methods that can be utilized to evaluate the performance of the model, improve the quality of predictions or present the structure of the model. It is a very wide category without distinctive representatives. Examples of explanation methods that comply with a definition of Model Diagnostics are residual density plots, variable vs prediction plots, and many other figures that present responses or residuals versus different variables. 

    \item The \textbf{Predict Parts} explanation task analyses and presents the impact of components and from the perspective of a single observation. The goal of that task is to measure the contributions each observation has to the final prediction. It belongs to the local explanation domain. Examples of the Predict Parts Explanation Method are LIME \citep{lime}, SHAP \citep{NIPS2017_7062} or BreakDown \citep{iBreakDown}.
    Probably the most popular method in this group is the SHapley Additive exPlanations (SHAP) \citep{NIPS2017_7062} method. It is based on the Shapley values concept, which derives from game theory. They are the solution for the problem in cooperative games where non-identical players contribute, to different extents, to the outcome. Shapley values allow us to calculate how the surplus should be distributed. Translating it into the language of models, players now are dependent variables while the model prediction is a surplus to divide. Current implementations usually approximate Shapley values in general case or calculate exact values for tree ensembles. 

    \item \textbf{Predict Profile} sometimes called sensitivity analysis, is a task similar to Model Profile. The only difference lies in the explanation subject. For this task, there is no aggregation over the whole dataset. Instead, the effects of variables are explained from the perspective of a single observation. An instance of an explanation method that complies with the Predict Profile task is Individual Conditional Expectations \citep{ICE} also known as the Ceteris Paribus Profile \citep{ema}. 
    
    \item \textbf{Predict Diagnostics} follows the same principia as the Model Diagnostics task. What differs between the two is the way they look at the model. For Predict Diagnostics, a single observation or its neighbors are taken as perspective instead of the whole dataset. One of the examples of the Predict Diagnostics methods is a comparison of residual distribution for $k$ neighbors of a~single observation versus distribution over the whole dataset.
\end{itemize}

\section{Comparison of packages for XAI analysis}

\subsection{Selection criteria for XAI packages}
\label{sec:criteria}

The number of tools that can be used to analyze predictive models is growing rapidly, and, at the same time, there is no established definition of an XAI tool. Therefore, it would not be possible to reliably identify and present all existing XAI packages. 

For a reliable analysis\footnote{The authors have made every effort to maintain an objective view when selecting and analyzing packages. However, a potential conflict of interest should be noted, we are authors or co-authors of some of the described packages, which belong to the \url{DrWhy.ai} family.}, we have adopted the following criteria. For the initial set of packages, we have used the list \textit{Awesome Machine Learning Interpretability}, maintained by Patrick Hall \citep{PatrickHall}. We chose this list because of the pioneer works related to the synthesis of approaches to XAI described in \textit{An Introduction to Machine Learning Interpretability} \citep{hall2018introduction,hall2018art}. 
We restricted this list to packages published on CRAN with at least 5000 downloads. The list is in Table~\ref{tab:pkgs_list}. 

In addition, this list was extended by packages available on CRAN, which have in the DESCRIPTION file some keywords used in XAI analysis. We checked 15,993 packages (access 08.07.2020) looking for keywords like \textit{xai}, \textit{iml}, \textit{explain} and \textit{interpretability}. This way, we encountered a great deal of false-positive entries (for instance, packages that explain the meaning of HTTP codes or indeed use XAI methods but to solve other, well-defined problems, not to actually explain the predictive models). This list was then manually cleaned and we ended up with 27 packages.

We would like to point out that \CRANpkg{mcr} \citep{mcr} and \CRANpkg{smbinning} \citep{smbinning} may look like they do not actually belong to the XAI area. However, they were included in Patrick Hall's original list, and we would consider it biased to arbitrarily remove packages despite meeting all the requirements of being included. Therefore, we did our best to extract the XAI related functionalities of those packages.

Note, the purpose of this paper is to compare packages that can be used for models created in R. However, we have decided to add also some Python libraries for context.% it would be incomplete and unfair if we would not mention any of the Python libraries. 

\begin{table}
\centering
\footnotesize
\begin{tabular}{|c|l|c|c|c|r|r|c|}
\hline
 & Package & License & \begin{tabular}[c]{@{}c@{}}Date of \\last update \end{tabular} & \begin{tabular}[c]{@{}c@{}}Last \\version \end{tabular} & \begin{tabular}[c]{@{}c@{}}GitHub \\stars\end{tabular}  & \begin{tabular}[c]{@{}c@{}}CRAN \\downloads\end{tabular} & \begin{tabular}[c]{@{}c@{}}Date of \\first release \end{tabular}\\
\hline
 & ALEPlot & GPL-2 & 2018-05-24 & 1.1 & - & 36,831 & 2017-11-13\\
\cline{2-8}
 & auditor & GPL-2/GPL-3 & 2020-05-28 & 1.3.0 & 54 & 26,757 & 2018-05-11\\
\cline{2-8}
 & DALEX & GPL-2/GPL-3 & 2021-03-20 & 2.2.0 & 796 & 99,688 & 2018-02-28\\
\cline{2-8}
 & DALEXtra & GPL-2/GPL-3 & 2020-09-07 & 2.0 & 40 & 13,554 & 2019-09-19\\
\cline{2-8}
 & EIX & GPL-2 & 2021-03-23 & 1.2.0 & 12 & 10,491 & 2019-05-31\\
\cline{2-8}
 & ExplainPrediction & GPL-3 & 2018-01-07 & 1.3.0 & - & 27,589 & 2015-09-07\\
\cline{2-8}
 & fairness & MIT + addons & 2020-11-19 & 1.2.0 & 21 & 10,369 & 2019-09-27\\
\cline{2-8}
 & fastshap & GPL-2/GPL-3 & 2020-02-02 & 0.0.5 & 46 & 26,820 & 2019-11-22\\
\cline{2-8}
 & flashlight & GPL-2/GPL-3 & 2021-02-13 & 0.7.5 & 9 & 14,312 & 2019-08-25\\
\cline{2-8}
 & forestmodel & GPL-2 & 2020-07-19 & 0.6.2 & 22 & 36,568 & 2015-11-26\\
\cline{2-8}
 & fscaret & GPL-2/GPL-3 & 2018-05-08 & 0.9.4.4 & - & 44,857 & 2013-06-13\\
\cline{2-8}
 & iBreakDown & GPL-3 & 2020-07-29 & 1.3.1 & 57 & 55,192 & 2019-04-04\\
\cline{2-8}
 & ICEbox & GPL-2/GPL-3 & 2017-07-13 & 1.1.2 & 32 & 38,117 & 2013-10-18\\
\cline{2-8}
 & iml & MIT + addons & 2020-09-24 & 0.10.1 & 389 & 126,471 & 2018-03-13\\
\cline{2-8}
 & ingredients & GPL-3 & 2021-02-05 & 2.0.1 & 31 & 58,454 & 2019-04-09\\
\cline{2-8}
 & lime & MIT + addons & 2021-02-24 & 0.5.2 & 438 & 110,090 & 2017-09-15\\
\cline{2-8}
 & live & MIT + addons & 2020-01-15 & 1.5.13 & 34 & 18,009 & 2018-04-03\\
\cline{2-8}
 & mcr & GPL-3 & 2014-02-12 & 1.2.1 & - & 39,120 & 2012-07-24\\
\cline{2-8}
 & modelDown & Apache License 2.0 & 2020-04-15 & 1.1 & 102 & 9,624 & 2019-06-15\\
\cline{2-8}
 & modelStudio & GPL-3 & 2021-01-07 & 2.1.1 & 163 & 15,943 & 2019-09-03\\
\cline{2-8}
 & pdp & GPL-2/GPL-3 & 2018-08-27 & 0.7.0 & 71 & 188,853 & 2016-09-02\\
\cline{2-8}
 & randomForestExplainer & GPL-2/GPL-3 & 2020-07-11 & 0.10.1 & 175 & 39,944 & 2017-07-15\\
\cline{2-8}
 & shapper & GPL-2/GPL-3 & 2020-08-28 & 0.1.3 & 47 & 15,615 & 2019-03-02\\
\cline{2-8}
 & smbinning & GPL-2/GPL-3 & 2019-04-01 & 0.9 & - & 275,329 & 2015-02-15\\
\cline{2-8}
 & survxai & GPL-2/GPL-3 & 2020-08-28 & 0.2.2 & 9 & 10,694 & 2018-08-24\\
\cline{2-8}
 & vip & GPL-2/GPL-3 & 2020-12-17 & 0.3.2 & 129 & 118,346 & 2018-06-15\\
\cline{2-8}
\multirow{-27}{*}{\centering\arraybackslash \centering\arraybackslash \rot{\rlap{R}}} & vivo & GPL-2 & 2020-09-07 & 0.2.1 & 14 & 9,292 & 2019-06-17\\
\cline{1-8}
 & aix360 & Apache 2.0 & 2020-10-28 & 0.2.1 & 787 & - & 2019-08-08\\
\cline{2-8}
 & eli5 & MIT & 2021-01-23 & 0.11.0 & 2323 & - & 2016-09-15\\
\cline{2-8}
 & interpret & MIT & 2021-01-20 & 0.2.4 & 3572 & - & 2019-05-04\\
\cline{2-8}
 & lime & BSD & 2020-06-26 & 0.2.0.1 & 8552 & - & 2016-03-24\\
\cline{2-8}
 & shap & MIT & 2021-03-03 & 0.39.0 & 12083 & - & 2016-12-01\\
\cline{2-8}
\multirow{-4}{*}{\centering\arraybackslash \centering\arraybackslash \rot{\rlap{Python}}} & skater & MIT & 2018-09-21 & 1.1.2 & 973 & - & 2017-05-23\\
\hline
\end{tabular}
\caption{List of packages that will be compared. The type of license date of the last CRAN release, number of GitHub stars, CRAN downloads, and date of the first release are also presented. A dash means that information was not accessible, meaning either the package does not have a GitHub repository or is not available on CRAN (it is a Python package). Downloads were acquired using \CRANpkg{deepdep} \citep{deepdep} package. Access 2021-03-23.}
\label{tab:pkgs_list}
\end{table}

%\begin{figure}[htb]
%  \centering
%  \includegraphics[scale=0.5]{downloads.png}
%  \caption{Cumulated downloads for 10 most popular compared XAI packages according to CRAN.}
%  \label{fig:downloads}
%\end{figure}

%To determinate causes of failure of our method we inspected description of packages that we kno  w are useful Explainable Machine Learning tools. It allowed us to understand what happened. Descriptions of R packages are very inconsistent. For instance, \CRANpkg{pdp} package \citep{pdp} has a very enigmatic description which is one sentence long and states that the purpose of this package is to create partial dependence plots for various machine learning models. On the other hand description of \CRANpkg{lime} package \citep{limeR} which was also made based on a single method, explains that methods shortly but also provides information that the purpose of this tool is to \textit{explain} prediction of the machine learning model. Further inspection of other descriptions confirmed our suspicions. They are inconsistent and sometimes it is hard or even impossible to categorize packages as XAI related based only on their description.

\clearpage

\subsection{Available methods in XAI packages}

\label{sec:packages_comaprison}

%XAI toolkits due to their huge diversity provide various way to provide methods that realize different explanation tasks. 

Table \ref{tab:pkgs_pyramid} summarizes methods available in the XAI packages.
Some packages are focused only on a~single aspect of a model explanation, while others implement a larger number of methods. In Section~\ref{sec:gallery} \textit{Example gallery for XAI packages}, we present an explanation generated by each package.

Most of the compared packages implement Model Parts or Predict Parts explanations. This means that the primary focus is on inspection of how consecutive variables contribute to model prediction either locally or globally. Nine out of all compared  R packages implement both Model Parts and Predict Parts methods. Two packages implement exactly those two types of methods, while seven provide other types of methods as well. On the other hand, Predict Diagnostics is the least represented type of explanation as only one package support it. However, Model Diagnostics methods are available in nine out of checked R packages, and, for three of them, this is the only supported type of explanation. Only five packages allow the variables to be locally profiled, while nine provide an opportunity to do it globally. Eight of the compared packages provide access to three or more different types of explanations, but only five to at least four types. Taking into consideration the compared Python packages, they give access to similar features as the R packages do.

\begin{table}[h!]
\centering
\small
\begin{tabular}{|c|c|c|c|c|c|c|c|} 
                          \multicolumn{2}{l}{\quad \quad \quad \quad}        & \multicolumn{3}{l}{\quad \quad \quad Global Explanations} & \multicolumn{3}{l}{\quad \quad \quad Local Explanations} \\ 
\hline
                                  & Package               & \begin{tabular}[c]{@{}c@{}}Model\\parts\end{tabular}      & \begin{tabular}[c]{@{}c@{}}Model\\profile\end{tabular}  & \begin{tabular}[c]{@{}c@{}}Model\\diagnostics\end{tabular} & \begin{tabular}[c]{@{}c@{}}Predict\\parts\end{tabular} & \begin{tabular}[c]{@{}c@{}}Predict\\profile\end{tabular} & \begin{tabular}[c]{@{}c@{}}Predict\\diagnostics\end{tabular} \\ 
\hline
\multirow{26}{*}{\rot{\rlap{R}}}               & ALEPlot               & -        & \cmark                                                  &   -     & -      & -  & -                                               \\ 
\cline{2-8}
                                  & auditor               & -   & -                                                   & \cmark       & - & -           & -                                     \\ 
\cline{2-8}
%                                  & DALEX                 & \cmark  & \cmark                                                   & \cmark & \cmark & \cmark         & \cmark                                              \\ 
\cline{2-8}
                                  & DALEX/DALEXtra            & \cmark   & \cmark                                                   & \cmark & \cmark & \cmark     & \cmark                                                \\ 
\cline{2-8}
                                  & EIX                   & \cmark         & -                                                   & \cmark & \cmark & -     & -                                                  \\ 
\cline{2-8}
                                  & ExplainPrediction     & \cmark        & -                                                   & - & \cmark & -      & -                                                \\ 
\cline{2-8}
                                  & fairness              & -  & -                                                   & \cmark  & - & -             & -                                          \\ 
\cline{2-8}
                                  & fastshap              & \cmark   & \cmark                                                    & - & \cmark & -          & -                                            \\ 
\cline{2-8}
                                  & flashlight            & \cmark  & \cmark                                                   & \cmark & \cmark & \cmark     & -                                                   \\ 
\cline{2-8}
                                  & forestmodel           & \cmark        &  -                                                  & - & - & -  & -                                                   \\ 
\cline{2-8}
                                  & fscaret               & \cmark  & -                                                   & - & - & -      & -                                                 \\ 
\cline{2-8}
%                                  & iBreakDown            & -  & -                                                   & - & \cmark & -         & -                                              \\ 
\cline{2-8}
                                  & ICEbox                & -   & \cmark                                                    & - & - & -        & -                                                \\ 
\cline{2-8}
                                  & iml                   & \cmark  & \cmark                                                    & - & \cmark & \cmark    & -                                                   \\ 
\cline{2-8}
%                                  & ingredients           & \cmark  & \cmark                                                   & - & \cmark & \cmark         & -                                              \\ 
\cline{2-8}
                                  & lime                  & -  & -                                                   & -    & \cmark  & - & -                                                  \\ 
\cline{2-8}
                                  & live                  & -  & -                                                   & -      & \cmark & -  & -                                                 \\ 
\cline{2-8}
                                  & mcr                   & -        & -                                                   & \cmark       & - & - & -                                                 \\ 
\cline{2-8}
                                  & modelDown             & \cmark    & \cmark                                                    & \cmark     & - & - & -                                                   \\ 
\cline{2-8}
                                  & modelStudio           & \cmark        & \cmark                                                    & -   & \cmark  & \cmark  & -                                                    \\ 
\cline{2-8}
                                  & pdp                   & \cmark   & \cmark                                                    & -     & - & - & -                                                  \\ 
\cline{2-8}
                                  & randomForestExplainer & \cmark    & -                                                   & -      & - & - & -                                                 \\ 
\cline{2-8}
                                  & shapper               & -   & -                                                   & -    & \cmark  & - & -                                                   \\ 
\cline{2-8}
                                  & smbinning             & \cmark   & -                                                   & \cmark      & - & - & -                                                 \\ 
\cline{2-8}
                                  & survxai               & \cmark   & -                                                   & \cmark    & \cmark  & \cmark  & -                                                    \\ 
\cline{2-8}
                                  & vip                   & \cmark   & -                                                   & -  & - & - & -                                                     \\ 
\cline{2-8}
                                  & vivo                   & \cmark   & -                                                   & -  & \cmark  & - & -                                                     \\ 
\hline
\multirow{8}{*}{\rot{\rlap{Python}}} & aix360                & \cmark & \cmark                                                  & \cmark    & \cmark & - & -                                                  \\ 
\cline{2-8}
                                  & eli5                  & \cmark           & -                                                   & -   & \cmark & - & -                                                    \\ 
\cline{2-8}
                                  & interpret             & \cmark         & \cmark                                                  & - & \cmark & - & -                                                      \\ 
\cline{2-8}
                                  & lime                  & -           & -                                                   & - & \cmark & - & -                                                      \\ 
\cline{2-8}
                                  & shap                  & \cmark          & \cmark                                                   & - & \cmark & - & -                                                      \\ 
\cline{2-8}
                                  & skater                & \cmark           & \cmark                                                   & -  & \cmark & - & -                                                     \\
\hline
\end{tabular}
\caption{Groups of methods (see Table~\ref{tab:hierarchy}) implemented in considered XAI packages. \cmark means that selected package implements at least one method that belongs to a given explanation task. As \pkg{DALEXtra} depends on \pkg{DALEX}, it inherits every functionality \pkg{DALEX} has. Package \pkg{ingredients} and \pkg{iBreakDown} were not included in this table since they are imported by \pkg{DALEX}, \pkg{modelStudio}  and \pkg{modelDown}. Access 2020-08-15.}
\label{tab:pkgs_pyramid}
\end{table}

\subsection{Models comparisons and the Rashomon effect}

\label{sec:rashomon}
 
The Rashomon effect \citep{rashomon} refers to a situation in which an event has contradictory interpretations by different spectators.  \cite{breiman2001statistical} has introduced this concept to machine learning modeling. Two or more models of similar performance may correspond to a different relation between variables. 

The juxtaposition of explanations for different models allows for a more complete analysis. However, not every package facilitates easy model comparison, and not every package was designed with model comparison in mind.
Packages \CRANpkg{auditor} \citep{auditor}, \pkg{DALEX}, \CRANpkg{DALEXtra} \citep{DALEXtra}, \CRANpkg{flashlight} \citep{flashlight}, \CRANpkg{ingredients} \citep{ingredients}, \CRANpkg{modelDown} \citep{modelDown}, \CRANpkg{modelStudio} \citep{modelStudio}, and \CRANpkg{vivo} \citep{vivo} are designed for plot explanations of two or more models next to each other in a single chart.

\subsection{Interoperability XAI frameworks}

\label{sec:interoperability}

Most of the packages we discuss in this article implement model-agnostic explanation methods. Yet, not all of them work smoothly with the different frameworks used to train predictive models. In Table~\ref{tab:pkgs_inter}, we summarize which packages work with popular machine learning frameworks. 
Such analysis is important because several models (for example random forest) are available in different frameworks (\pkg{mlr}, \pkg{scikit-learn}, and others) and although algorithms should be the same, the models differ in terms of names of parameters or interfaces to making predictions. Therefore, to explain models created with various ML frameworks it may be necessary to put additional effort into obtaining the prediction of the model in the form understandable for the XAI framework.

Interoperability with the framework is not binary, and it may require a different amount of work. The packages can support the given framework out of the box, meaning they do not require any additional work to generate an explanation except providing a model or loading any additional library. Packages may also allow for relatively easy use of frameworks, for example, bypassing their own user-defined function that accesses model predictions. If the explanations and model are in two different programming languages, the application of one to another may require even more work, e.g. using the \CRANpkg{reticulate} \citep{reticulate} package.

Overall, we compared 27 R packages and six Python libraries in terms of their interoperability with various machine learning frameworks. Four of those frameworks are implemented in R, two of them in Python and one in Java, which is accessible via official R and Python wrappers. Four compared XAI packages, \CRANpkg{EIX} \citep{EIX}, \CRANpkg{forestmodel} \citep{forestmodel}, \CRANpkg{randomForestExplainer} \citep{randomForestExplainer} and \pkg{smbinning}, present a model-specific approach and, are designed to work only with a particular group of models. One R package, \pkg{mcr}, does not allow model input at all. Instead, it fits and explains its own algorithms. Only one package (i.e. \CRANpkg{fscaret} \citep{fscaret}) supports one framework despite implementing model-agnostic methods, which is due to the fact that it is an extension for the \pkg{caret} ML framework. To the best of our knowledge, two of the compared packages, \pkg{DALEXtra} and \pkg{modelStudio} provide support for all types of models taken into consideration. Moreover, there is no universal ML framework that would be supported by all XAI packages. The \pkg{caret} package is supported by the highest number of packages, while, on the other hand, \pkg{mlr3} by the least, which is probably related to the time of development of these frameworks.

\subsection{Time of operation}

The time of operation is an important aspect of software and varies between different XAI packages. Some explanations are generated in near real-time, while others take a long time to produce results. In Table~\ref{tab:benchmark}, we have an overview of computation time for different XAI packages.
We performed this benchmark to enhance the comparison of R packages. The benchmark was performed on a standard laptop used for everyday work. We computed the time of evaluation of each chunk from Markdown files linked in Section~\ref{sec:gallery}.

\begin{table}[H]
\centering
\small
\begin{tabular}{|c|l|c|c|c|c|c|c|c|} 
                          \multicolumn{2}{l}{\quad \quad \quad \quad}        & \multicolumn{4}{l}{\quad \quad \quad \quad \quad \quad \quad R} & \multicolumn{2}{l}{\quad \quad Python} & \multicolumn{1}{l}{Java} \\ 
\hline
                                  & Package               & mlr      & mlr3  & parsnip & caret & keras & scikit-learn & h2o \\ 
\hline
\multirow{26}{*}{\rot{\rlap{R}}}               & ALEPlot               & $\star$        & $\star$                                                  &   $\star$     & $\star$        & $\bullet$  & $\bullet$ & $\star$                                              \\ 
\cline{2-9}
                                  & auditor               & $\star$   & $\star$                                                   & \cmark       & \cmark    & $\bullet$  & $\bullet$   & $\star$                                \\ 
\cline{2-9}
                                  & DALEX                 & $\star$   & $\star$                                                   & \cmark       & \cmark    & $\bullet$  & $\bullet$   & $\star$                                \\ 
\cline{2-9}
                                  & DALEXtra             & \cmark   & \cmark                                                   & \cmark & \cmark & \cmark     & \cmark  & \cmark                                           \\ 
\cline{2-9}
                                  & EIX\tablefootnote{\pkg{EIX} is a package that provide model-specific explanations of packages created with xgboost or \mbox{\pkg{LightGBM}~packages}.}                   & -         & -                                                      & -          & -      & -  & -  & -                                          \\ 
\cline{2-9}
                                  & ExplainPrediction     & $\star$        & $\star$                                                    & $\star$  & $\star$        & $\bullet$   & $\bullet$      & $\star$                                        \\ 
\cline{2-9}
                                  & fairness              & $\star$   & $\star$                                                                & $\star$       & $\star$       & $\bullet$   & $\bullet$ & $\star$                                       \\ 
\cline{2-9}
                                  & fastshap              & $\star$   & $\star$                                                   & $\star$ & $\star$     & $\bullet$  & $\bullet$         & $\star$                              \\ 
\cline{2-9}
                                  & flashlight            & $\star$ & $\star$                                                   & $\star$ & $\star$       & $\bullet$  & $\bullet$      & $\star$                                          \\ 
\cline{2-9}
                                  & forestmodel\tablefootnote{\pkg{forestmodel} is a package that provide model-specific explanations of linear models created with stats package}            & -       &  -                                                  & - & -    & -  & -  & -                                                  \\ 
\cline{2-9}
                                  & fscaret               & -  & -                                                   & - & \cmark       & -   & -  & -                                                \\ 
\cline{2-9}
                                  & iBreakDown           & $\star$   & $\star$                                                   & \cmark       & \cmark    & $\bullet$  & $\bullet$   & $\star$                                \\ 
\cline{2-9}
                                  & ICEbox                & $\star$    & $\star$                                                     & $\star$  & $\star$    & $\bullet$  & $\bullet$       & $\star$                                           \\ 
\cline{2-9}
                                  & iml                   & \cmark  & $\star$                                                   & $\star$ & \cmark   & $\bullet$   & $\bullet$             & $\star$                                       \\ 
\cline{2-9}
                                  & ingredients           & $\star$   & $\star$                                                   & \cmark       & \cmark    & $\bullet$  & $\bullet$   & $\star$                                \\ 
\cline{2-9}
                                  & lime                  & \cmark  & $\star$                                                   & \cmark     & \cmark    & $\bullet$  & $\bullet$  & \cmark                                             \\ 
\cline{2-9}
                                  & live                  & $\star$  & $\star$                                                  & $\star$      & $\star$    & $\bullet$  & $\bullet$  & $\star$                                                \\ 
\cline{2-9}
                                  & mcr\tablefootnote{\pkg{mcr} package build its own model and present explanation for them.}                    & - & - & -      & -      & -  & -   & -                                             \\ 
\cline{2-9}
                                  & modelDown             & $\star$   & $\star$                                                   & \cmark       & \cmark    & $\bullet$  & $\bullet$   & $\star$  \\
\cline{2-9}
                                  & modelStudio            & \cmark   & \cmark                                                   & \cmark & \cmark & \cmark     & \cmark  & \cmark                                           \\ 
\cline{2-9}
                                  & pdp                   & $\star$   & $\star$                                                    & $\star$     & $\star$     & $\bullet$  & $\bullet$   & $\star$                                               \\ 
\cline{2-9}
                                  & randomForestExplainer\tablefootnote{\pkg{randomForestExplainer} is a package that provide model-specific explanations of forest models created with \pkg{randomForest} package} & -    & -                                                   & -      & -  & -  & -      & -                                           \\ 
\cline{2-9}
                                  & shapper               & $\star$   & $\star$                                                   & $\star$   & $\star$    & $\bullet$  & $\bullet$  & $\star$                                                \\ 
\cline{2-9}
                                  & smbinning\tablefootnote{\pkg{smbinning} is a package that provide model-specific explanations of scoring linear models created among others with stats package}             & -  & -                                                   & -      & -  & -  & -  & -                                             \\ 
\cline{2-9}
                                  & survxai               & $\star$  & $\star$                                                   & $\star$   & -     & -  & - & -                                                 \\ 
\cline{2-9}
                                  & vip                   & $\star$   & $\star$                                                  & \cmark  &\cmark   & $\bullet$  & $\bullet$     & $\star$                                        \\ 
\cline{2-9}
                                  & vivo                    & $\star$   & $\star$                                                   & \cmark       & \cmark   & $\bullet$  & $\bullet$   & $\star$  \\ 
\hline
\multirow{8}{*}{\rot{\rlap{Python}}} & aix360\tablefootnote{\pkg{aix360} package build its own model and present explanation for them.}         & - & - & -   & -      & -  & \cmark  & -                                                 \\ 
\cline{2-9}
                                  & eli5                  & -           & -                                                   & -     & -      & \cmark  & \cmark  & -                                                  \\ 
\cline{2-9}
                                  & interpret\tablefootnote{\pkg{interpret} package build its own model and present explanation for them.}              & -        & -                                          & - & -      & -  & \cmark  & -                                                      \\ 
\cline{2-9}
                                  & lime                  & -           & -                                                   & - & -      & \cmark   & \cmark  & $\star$                                                     \\ 
\cline{2-9}
                                  & shap                  & -          & -                                                  & - & -       & \cmark  & \cmark   & $\star$                                                 \\ 
\cline{2-9}
                                  & skater                & $\bullet$           & $\bullet$                                                  & $\bullet$  & $\bullet$    & \cmark   & \cmark & $\star$                                                    \\
\hline
\end{tabular}
\caption{Interoperability of XAI toolkits. \cmark stands for the support that the XAI framework gives for the modeling framework. Support means that computing explanations do not require any additional steps besides providing a model. $\star$ stands for partial support, which means that it is necessary to pass additional user-defined functions that extracts models' predictions. For instance, R~package \pkg{flashlight}, requires passing two-argument function \texttt{function(model, newdata)} that returns prediction vector.  $\bullet$ mark symbolizes that obtaining predictions requires greater effort, for instance, manual configuration of R-Python connection via \pkg{reticulate}. Keep in mind that marks in the h2o column refer to interoperability with R or Python h2o ports.}
\label{tab:pkgs_inter}
\end{table}

\begin{table}[ht]
\centering
\footnotesize
\begin{tabular}{|p{2cm}|r|r|r|r|r|r|r|r|r|}
  \hline
 Package & \makecell{Model \\ parts} & \makecell{Model \\ profile} & \makecell{Model \\ diagnostics} & \makecell{Predict \\ parts} & \makecell{Predict \\ profile} & \makecell{Predict \\ diagnostics} & Report & Sum & N \\ 
  \hline
ALEplot & - & 0.29 & - & - & - & - & - & 0.58 &   2 \\ 
\hline
  auditor & - & - & 0.18 & - & - & - & - & 65.88 &   9 \\ 
  \hline
  DALEX & 6.01 & 0.52 & 0.22 & 1.81 & 0.49 & 0.41 & - & 36.18 &  18 \\ 
  \hline
  DALEXtra & - & - & 0.42 & - & - & - & 9.15 & 11.44 &   4 \\ 
  \hline
  EIX & 5.15 & - & 5.14 & 26.58 & - & - & - & 42.02 &   4 \\ 
  \hline
  ExplainPrediction & 38.25 & - & - & 38.87 & - & - & - & 77.12 &   2 \\ 
  \hline
  fairness & - & - & 0.1 & - & - & - & - & 0.20 &   2 \\ 
  \hline
  fastshap & 107.92 & 106.3 & - & 107.93 & - & - & - & 430.09 &   4 \\ \hline
  flashlight & 7.44 & 0.17 & - & 0.8 & 0.17 & - & - & 50.62 &  19 \\ \hline
  fscaret & 74.06 & - & - & - & - & - & - & 74.06 &   1 \\ 
  \hline
  iBreakDown & - & - & - & 3.55 & - & - & - & 25.65 &   3 \\ 
  \hline
  ICEbox & - & 31.36 & - & - & - & - & - & 91.54 &   3 \\ 
  \hline
  iml & 42.3 & 0.27 & - & 0.42 & 2.54 & - & - & 115.02 &   9 \\ 
  \hline
  ingredients & 9.8 & 1.35 & - & 0.25 & 0.06 & - & - & 142.21 &  10 \\ 
  \hline
  lime & - & - & - & 0.4 & - & - & - & 0.80 &   2 \\ 
  \hline
  live & - & - & - & 2.5 & - & - & - & 5.00 &   2 \\ 
  \hline
  mcr & - & - & 0.48 & - & - & - & - & 2.18 &   6 \\ 
  \hline
  modelDown & - & - & - & - & - & - & 62.98 & 125.95 &   2 \\ 
  \hline
  modelStudio & - & - & - & - & - & - & 183.09 & 183.09 &   1 \\ 
  \hline
  pdp & 0.22 & 0.06 & - & - & - & - & - & 0.45 &   4 \\ 
  \hline
  randomForest- Explainer & 44.47 & - & - & - & - & - & 478.8 & 1027.16 &   7 \\ 
  \hline
  shapper & - & - & - & 4.95 & - & - & - & 9.90 &   2 \\ 
  \hline
  smbinning & - & - & 1.41 & - & - & - & - & 3.85 &   3 \\ 
  \hline
  survxai & - & 0.85 & 0.03 & - & 0.13 & - & - & 1.86 &   4 \\ 
  \hline
  vip & 1.69 & - & - & - & - & - & - & 310.58 &   5 \\ 
  \hline
  vivo & 1.34 & - & - & 1.22 & - & - & - & 2.56 &   2 \\ 
   \hline
\end{tabular}
\caption{Summarized times of computations (measured in seconds) for functions used in R markdowns linked in Section~\ref{sec:gallery}. Columns that correspond to types of explanations contain median time over chunks. Column Report contains a median time of report generation. Column Sum contains the overall time of computation of all chunks. Column N contains the number of chunks that were evaluated. Dashes mean that the package does not implement such an explanation type. There might be some differences between this Table and Table~\ref{tab:pkgs_pyramid} that comes from the design of packages; for example, \pkg{modelStudio} provides Model Profile explanations, yet only inside a report.}
\label{tab:benchmark}
\end{table}

Wherever it was possible, we tried to be consistent with the parameters of functions from different packages. For example, the number of samples for Shapley-based method $= 50$. However, given that markdowns contain only sample codes, the default parameters between functions in different packages may cause differences in code evaluation times. Therefore, the benchmark should not be considered the final comparison of the speed of the packages, but rather the support in developing an~intuition for the overall performance time. 

The packages \pkg{DALEXtra}, \pkg{modelStudio}, \pkg{modelDown}, and \pkg{randomForestExplainer} have long computation times, however, they are designed to compute standalone reports that can be later viewed without any additional computations. XAI frameworks, such as, \pkg{DALEX}, \pkg{flashlight}, and \mbox{\CRANpkg{iml}~\citep{iml}~have} a~wide range of computation times because of the several different methods they implement.
It is worth noting that we calculated all times for the models fitted to the titanic data set that has 2,207 observations, which is a relatively small amount of data. For large data sets, some explanations may take a long time or be impossible to compute. 

\section{Recommendations for a gentle introduction to XAI}
\label{sec:recommendation}

In this Section, we provide recommendations on how to begin explaining models.  We covered the most popular methods, such as permutational variable importance, SHAP explanation, Partial Dependence Profiles, ICE profile, and fairness check. We briefly describe the methods and their important parameters. Moreover, we provide code snippets with examples from the frameworks that we believe are best for getting started with XAI, i.e. \pkg{DALEX}, \pkg{flashlight}, and \pkg{iml}. 
% The only exception is fairness check, for which we show examples for packages \pkg{fairmodels} and \pkg{fariness}.
All examples in this Section are based on a  \CRANpkg{ranger} \citep{ranger} model trained on the titanic data set from the \pkg{DALEX}~package. 
\begin{example*}
data(titanic_imputed, package = "DALEX")
ranger_model <- ranger::ranger(survived~., data = titanic_imputed, 
                               classification = TRUE, probability = TRUE)
\end{example*}

Different libraries in R generate models whose predict function returns data in different formats. Model explanation tools need a unified interface for calculating predictions. The following function is such a unifying API.  It takes a ranger model and data, then returns a score vector as the result. For a~\verb'ranger' object, this requires pulling a second column from the \verb'predictions' slot.

\begin{example*}
flashlight_predict <- function(X.model, new_data) 
        predict(X.model, new_data)\$predictions[,2]
iml_predict <- function(X.model, newdata) 
        predict(X.model, newdata)\$predictions[,2]
\end{example*}

\newpage

The commonly used XAI technique is the model agnostic assessment of the importance of variables. Permutational variable importance \citep{fi} is a popular choice since it works in a model agnostic fashion. We repeatedly permute the values of a variable and examine how the performance of a model changes on average. The more the model's performance decreases, the more important the variable is.

The two most important parameters of this method are the number of iterations and the loss function. More iterations increase the stability of estimation, at the cost of longer execution time. 
Another important parameter for this method is the performance measure. Note that sometimes a loss function is used for this purpose.

In each of packages \pkg{DALEX}, \pkg{flashlight}, and \pkg{iml} it is possible to specify both the number of permutations (parameters \verb'B', \verb'm_repetitions', and \verb'n.repetitions', respectively) and the loss function (\verb'loss_function', \verb'metrics', and \verb'loss'). In each case one can specify also a custom loss function or a~function from another package, like \CRANpkg{MetricsWeighted}  \citep{MetricsWeighted}. 
 
% The only difference between compared packages is that \pkg{iml} package that does not support AUC as the measure by default.  
Figure~\ref{fig:fi_rec} shows the permutational variable importance generated by each of the packages. To make the results as comparable as possible, we tried to keep the same values of parameters for \mbox{each implementation.}

  \begin{minipage}{\linewidth}
      \centering
      \begin{minipage}{0.45\linewidth}
          \begin{figure}[H]
              \includegraphics[width=0.9\linewidth]{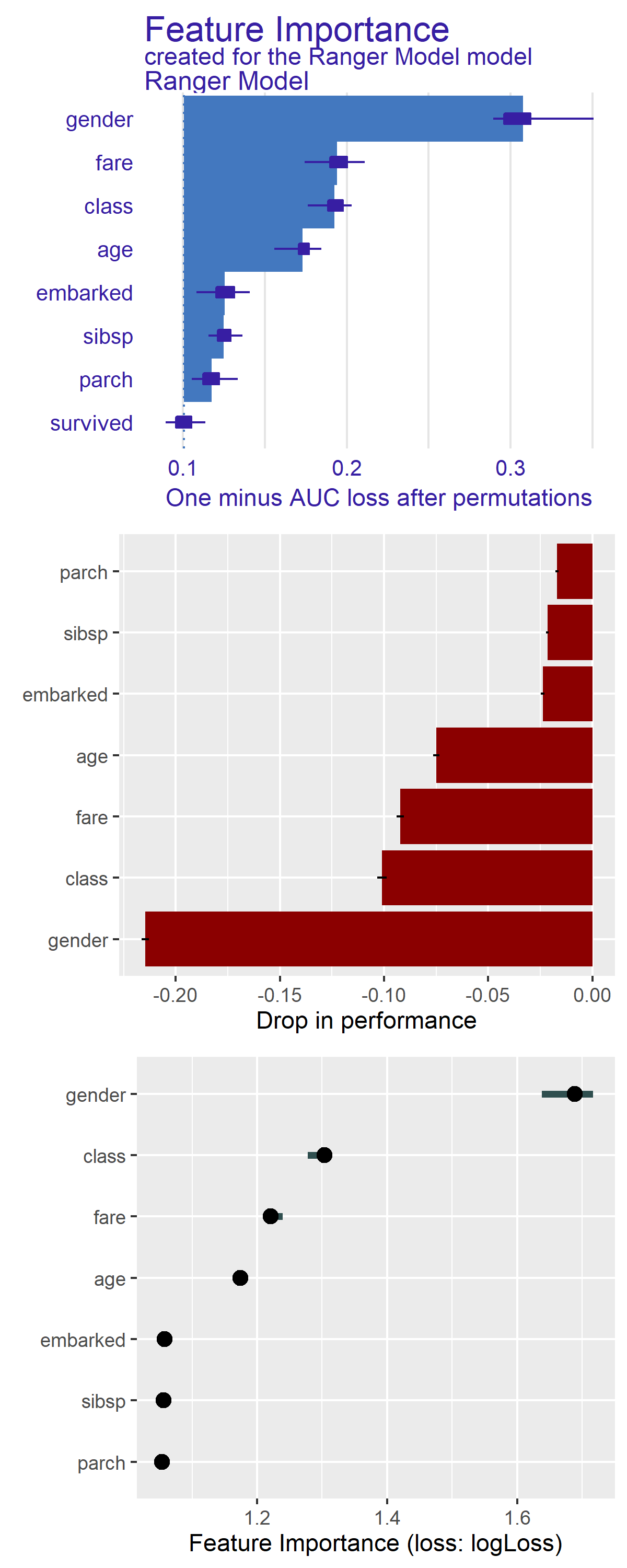}
              \caption{Permutational variable importance examples for different libraries. From the top there are \pkg{DALEX}, \pkg{flashlight} and \pkg{iml}.  The computation times of the plots are 4.59s for \pkg{DALEX}, 5.64s for \pkg{flashlight}, 8.86s for \pkg{iml}.}
              \label{fig:fi_rec}
          \end{figure}
      \end{minipage}
      \hspace{0.05\linewidth}
      \begin{minipage}{0.45\linewidth}
        \begin{example*}
library("DALEX")
exp_dalex <- 
    explain(ranger_model,
    data  = titanic_imputed,
    y     = titanic_imputed\$survived,
    label = "Ranger Model")
    
fi_dalex <- model_parts(exp_dalex, B = 10,
    loss_function = loss_one_minus_auc)
plot(fi_dalex)
        \end{example*}

                \begin{example*}
library("flashlight")
library("MetricsWeighted")
exp_flashlight <- 
    flashlight(model = ranger_model, 
    data    = titanic_imputed, 
    y       = "survived", 
    label   = "Titanic Ranger",
    metrics = list(auc = MetricsWeighted::AUC), 
    predict_function = flashlight\_predict)
    
fi_flashlight <- light_importance(exp_flashlight, 
    m_repetitions = 10)
plot(fi_flashlight, fill = "darkred")
        \end{example*}
                        \begin{example*}
library("iml")
X <- titanic_imputed[
which(names(titanic_imputed) != "survived")
]
exp_iml <- 
    Predictor\$new(ranger_model, 
    data = X, 
    y = titanic\_imputed\$survived, 
    predict.function = iml_predict)

fi_iml <- FeatureImp\$new(exp_iml, 
    loss = "logLoss",
    n.repetitions = 10)
plot(fi_iml)

        \end{example*}
      \end{minipage}
  \end{minipage}

\newpage

For the local prediction level exploration, the most popular XAI method for variable attribution is SHAP \citep{NIPS2017_7062}. The method uses Shapley value concept from game theory to estimate variables' contributions to a model's prediction for one observation. 
 
Shapley values are approximated with Monte-Carlo sampling. The approximation is based on a~number of random variables' orderings. The higher the number, the more stable is the approximation. Of course, the higher number of steps, the longer computations are. 

In each of the packages \pkg{DALEX}, \pkg{flashlight}, and \pkg{iml} one can specify the number of orderings (parameters \verb'B', \verb'n_perm', and \verb'sample.size', respectively). Additionally for \pkg{DALEX} and \pkg{flashlight} number of samples taken into consideration during computing the explanation is also an important parameter as it significantly affects computation time (\verb'N' and \verb'n_max' accordingly). This does not apply to \pkg{iml} due to the difference in the method chosen to calculate Shapley values.

Figure~\ref{fig:shap_rec} shows examples of SHAP explanation, note that once the number of orderings was set to the same value across all three packages, then the results are very similar. In this example \pkg{DALEX} and \pkg{flashlight} are slower than \pkg{iml}, in part due to the storing of partial results from individual permutations which makes it possible to determine error boxplots and in part because in each permutation \pkg{iml} takes into account a single observation per variable while \pkg{DALEX} and \pkg{flashlight} takes into account \verb'N'~/~\verb'n_max' observations. This is why results for \pkg{DALEX} and \pkg{flashlight} are in this example more stable than \pkg{iml} for the same number of permutations.  Note that the fastest implementation of SHAP values in R is available in the \pkg{treeshap} \citep{treeshap} package. 

% library("DALEX")
%explainer_dalex <- 
%    explain(ranger_model, 
%    data = titanic_imputed, y = titanic_imputed\$survived, 
%    label = "Ranger Model", 
%    verbose = FALSE)

%library("flashlight")
%custom_predict <- function(X.model, new_data) \{
%  predict(X.model, new_data)\$predictions[,2]
%\}
%explainer_flashlight <- 
%    flashlight(model = ranger_model, 
%    data = titanic_imputed, y = "survived", 
%    label = "Titanic Ranger",
%    metrics = list(auc = AUC), 
%    predict_function = custom\_predict)

%library("iml")
%custom_predict <- function(X.model, newdata) \{
%  predict(X.model, newdata)\$predictions[,2]
%\}
%X <- titanic_imputed[
%which(names(titanic_imputed) != "survived")
%]
%explainer_iml <- 
%    Predictor\$new(ranger_model, 
%    data = X, y = titanic\_imputed\$survived, 
%    predict.function = custom_predict)

   \begin{minipage}{\linewidth}
      \centering
      \begin{minipage}{0.45\linewidth}
          \begin{figure}[H]
              \includegraphics[width=0.85\linewidth]{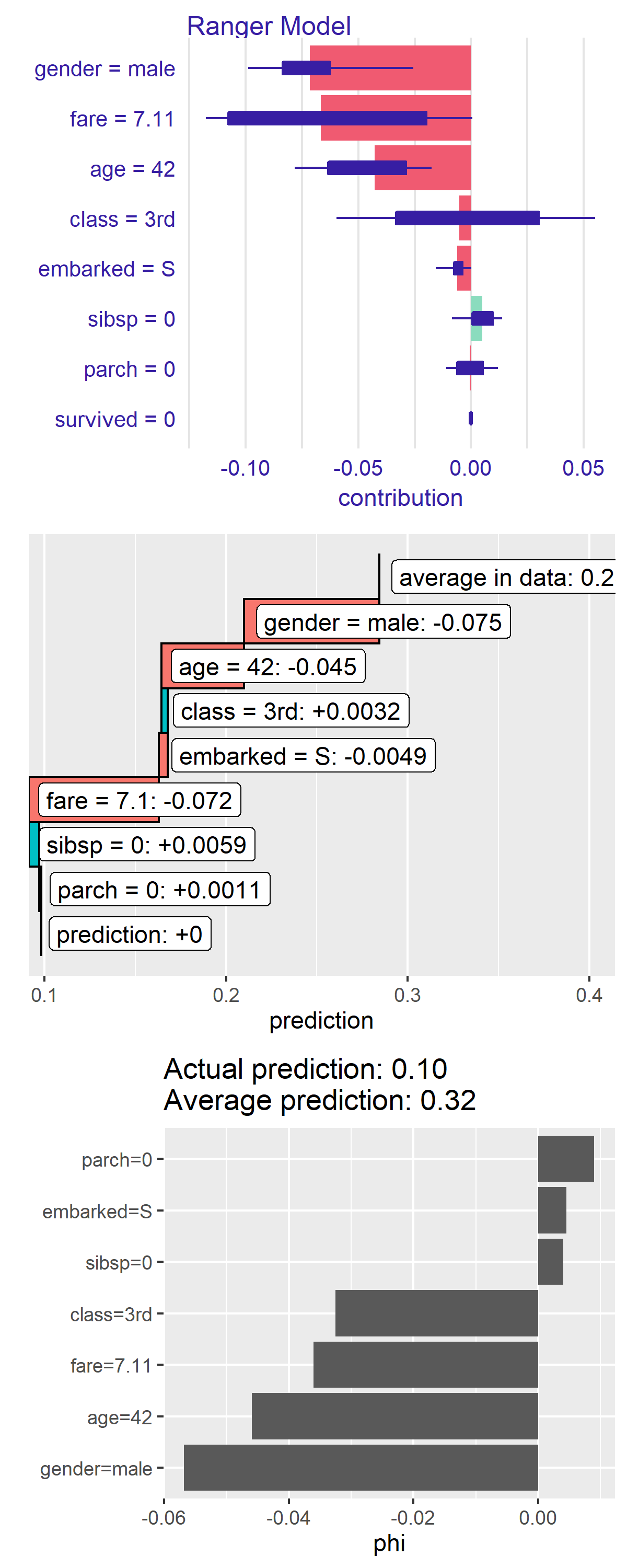}
              \caption{SHAP examples for different libraries. From the top there are \pkg{DALEX}, \pkg{flashlight} and \pkg{iml}. The computation times of the plots are 10.23s for \pkg{DALEX}, 7.11s for \pkg{flashlight}, 0.33s for \pkg{iml}.}
              \label{fig:shap_rec}
          \end{figure}
      \end{minipage}
      \hspace{0.05\linewidth}
      \begin{minipage}{0.45\linewidth}
        \begin{example*}
# DALEX
shap_dalex <- 
    predict_parts(exp_dalex, 
      new_observation = titanic_imputed[1,], 
      type = "shap",
      N    = 50, 
      B    = 50)
plot(shap_dalex)

        \end{example*}

                \begin{example*}
# flashlight
shap_flashlight <- 
    light_breakdown(exp_flashlight, 
      new_obs = titanic_imputed[1, ], 
      n_max   = 50,  
      n_perm  = 50,
      visit_strategy = "permutation"
   )
plot(shap_flashlight)

        \end{example*}
                        \begin{example*}
# iml
shap_iml <- 
    Shapley\$new(exp_iml, 
      x.interest  = X[1, ], 
      sample.size = 50)
plot(shap_iml)


        \end{example*}
      \end{minipage}
  \end{minipage}
  
\newpage

    Ceteris Paribus (CP) Profiles, known also as ICE curves \citep{ICE}, are profiles of the variables from the perspective of a single observation. CP profile shows how a model’s prediction would change if the value of a single variable changed and the other variables are fixed.
    
    The most important parameter of this method is the number of grid points (different values of a~variable) based on which the profile is constructed. It is also important for which observation we are calculating CP, since each observation has a different profile.
    
    In \pkg{DALEX}, \pkg{flashlight}, and \pkg{iml} it is possible to specify both the number of grid points (\verb'grid_points', \verb'n_bins', and \verb'grid.size' respectively). \pkg{DALEX} and \pkg{flashlight} allow to specify one observation for which profile shall be calculated (\verb'new_observation' and \verb'indices' accordingly) while in \pkg{iml} packages it is not possible and requires change of the reference data set.
    
    Figure~\ref{fig:cp_rec} shows an example of Ceteris Paribus profiles calculated using \pkg{DALEX}, \pkg{flashlight} and \pkg{iml} packages. To make results as comparable as possible, we tried to keep the same values of parameters for \mbox{each implementation.} The longer running time for \pkg{iml} is due to the fact that it calculates curves for each observation.

       \begin{minipage}{\linewidth}
      \centering
      \begin{minipage}{0.45\linewidth}
          \begin{figure}[H]
              \includegraphics[width=0.9\linewidth]{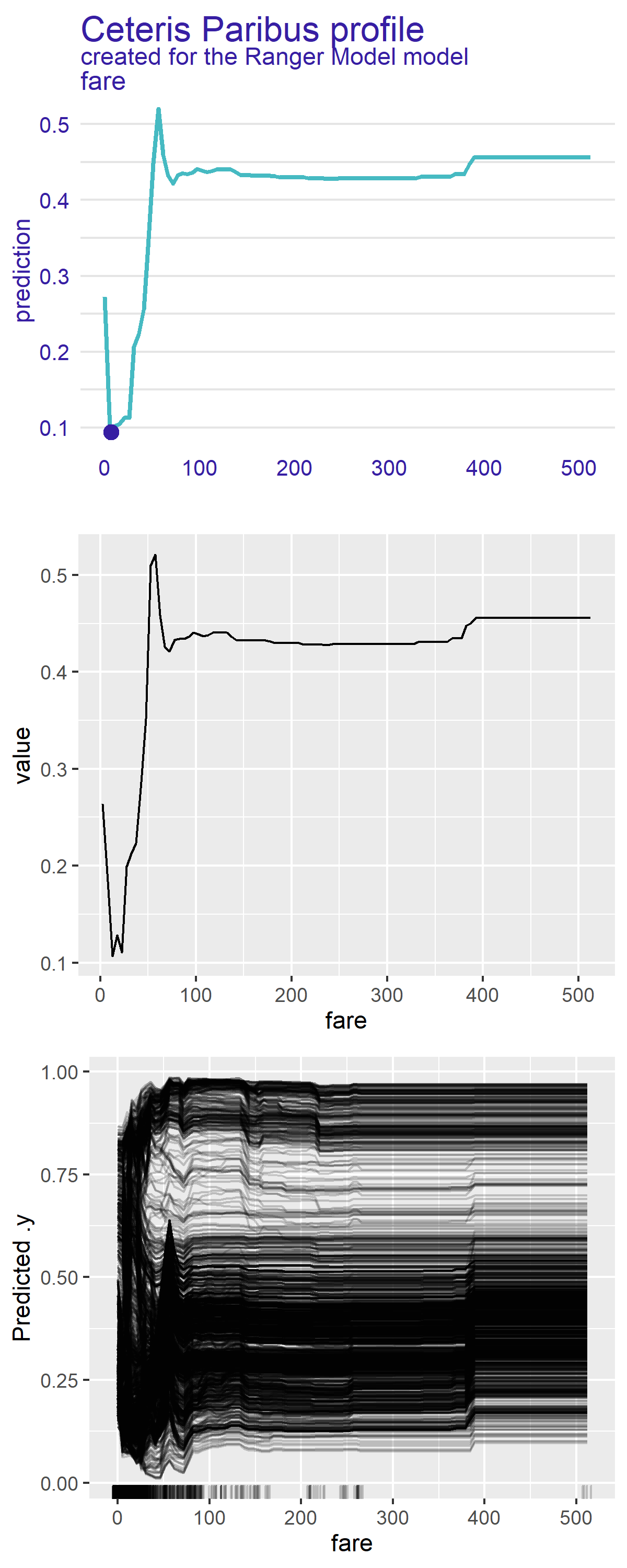}
              \caption{Ceteris Paribus Profiles examples for different libraries. From the top there are \pkg{DALEX}, \pkg{flashlight} and \pkg{iml}. The computation times of the plots are 0.05s for \pkg{DALEX}, 0.03s for \pkg{flashlight}, 10.02s for \pkg{iml}.}
              \label{fig:cp_rec}
          \end{figure}
      \end{minipage}
      \hspace{0.05\linewidth}
      \begin{minipage}{0.45\linewidth}
        \begin{example*}
# DALEX
cp_dalex <- 
    predict_profile(exp_dalex, 
         new_observation = titanic_imputed[1,], 
      variables       = "fare", 
      grid_points     = 101)
plot(cp_dalex, variables = "fare")

        \end{example*}

                \begin{example*}
# flashlight 
cp_flashlight <- 
    light_ice(exp_flashlight, 
      v       = "fare", 
      indices = 1,
      n_bins  = 101)
plot(cp_flashlight)

        \end{example*}
                        \begin{example*}
# iml
cp_iml <- 
    FeatureEffect\$new(exp_iml, 
      feature   = "fare", 
      method    = "ice",
      grid.size = 101)
plot(cp_iml)







        \end{example*}
      \end{minipage}
  \end{minipage}

\newpage

Partial Dependence Profile (PDP) is a common XAI technique that is useful to examine variables from a model perspective. It shows the global relationship between the dependent variable and model response by averaging Ceteris Paribus profiles for a chosen variable. Therefore, PDP is the profile of mean prediction that is a function of the independent variable.

First out of the two most important parameters is the number of grid points (values of variables) based on which the profile is constructed. The second is the distribution of the before-mentioned grid points. The PDP will be different, depending on whether we distribute the points uniformly or according to the empirical distribution of the variable.

Grid number of grid points can be changed in all of the three packages (\verb'grid_points' for \pkg{DALEX}, \verb'n_bins' for \pkg{flashlight} and \verb'grid.size' for \pkg{iml}). Yet, the grid distribution can be changed only in the \pkg{DALEX} and \pkg{flashlight} packages (parameter \verb'variable_splits_type' and \verb'cut_type').
Moreover, only \pkg{DALEX} and \pkg{flashlight} allow to set the number of observations used to construct PDP (\verb'N' and \verb'pd_n_max' accordingly). 

Examples of PDP curves in different packages are in Figure~\ref{fig:pdp_rec}. Although we tried to keep the parameters of the method the same, small differences are caused by different observation samples and grid distribution.
  
     \begin{minipage}{\linewidth}
      \centering
      \begin{minipage}{0.45\linewidth}
          \begin{figure}[H]
              \includegraphics[width=0.9\linewidth]{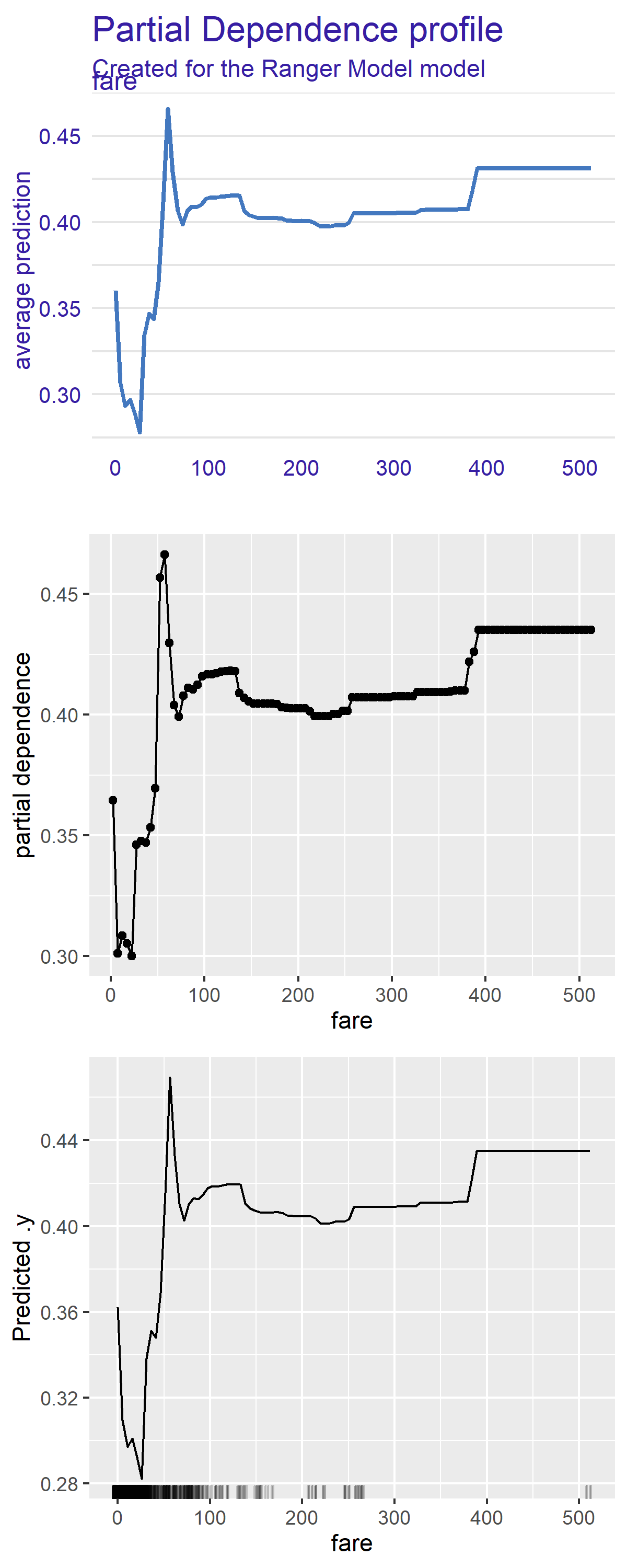}
              \caption{Partial Dependence Profiles examples for different libraries. From the top there are \pkg{DALEX}, \pkg{flashlight} and \pkg{iml}. The computation times of the plots are 2.99s for \pkg{DALEX}, 3.18s for \pkg{flashlight}, 11.28 s for \pkg{iml}.}
              \label{fig:pdp_rec}
          \end{figure}
      \end{minipage}
      \hspace{0.05\linewidth}
      \begin{minipage}{0.45\linewidth}
        \begin{example*}
# DALEX
pdp_dalex <-
    model_profile(
      exp_dalex, 
      variables   = "fare", 
      type        = "partial", 
      N           = 1000, 
      grid_points = 101, 
      variable_splits_type = "uniform")
plot(pdp_dalex)

        \end{example*}

                \begin{example*}
# flashlight
pdp_flashlight <- 
    light_profile(exp_flashlight, 
      v        = "fare", 
      type     = "partial dependence", 
      pd_n_max = 1000, 
      n_bins   = 101,
      cut_type = "equal")
plot(pdp_flashlight)

        \end{example*}
                        \begin{example*}
# iml
pdp_iml <- 
    FeatureEffect\$new(exp_iml, 
      feature   = "fare", 
      method    = "pdp", 
      grid.size = 101)
plot(pdp_iml)






        \end{example*}
      \end{minipage}
  \end{minipage}

\subsection{Unique features of selected XAI packages}

In the previous Section, we have covered popular and basic XAI techniques with examples from the frameworks, such as \pkg{DALEX}, \pkg{flashlight}, and \pkg{iml}. In this Section, we present general recommendations for those three packages based on our experience in XAI.
For key functionalities, each of these packages offers similar methods that can be used in similar ways. However, these packages differ in additional specialized functionality.

The \pkg{DALEX} package provides a unified wrapper for machine learning models that can be later utilized by other XAI packages. A rich system of additional packages allows many non-standard applications. For example, the \pkg{modelDown} package allows to automatically convert an explainer into static \href{http://xai-tools.drwhy.ai/modelDown/index.html}{html documentation} based on \verb'pkgdown' templates, which can be versioned or shared with other users. The \href{https://modeloriented.github.io/xai2shiny/}{xai2shiny} \citep{xai2shiny} package allows to automatically convert an explainer with just one line of code into a shiny application which allows for dynamic model exploration. The \pkg{modelStudio} and \CRANpkg{arenar} \citep{arenar} packages allow you to build, with a single command, an interactive \href{http://xai-tools.drwhy.ai/modelStudio_titanic.html}{java script} based tool for model exploration using the IEMA approach. The \CRANpkg{fairmodels} \citep{fairmodels} package does fairness analysis of the model for similar behavior in subpopulations. 
The \pkg{triplot} \citep{triplot}  package allows you to calculate the importance of not only individual variables but also whole aspects of variables. The \pkg{auditor}  package allows very detailed analysis of model residuals while the \CRANpkg{drifter} package allows analysis of drift in the data and in the model structure. In addition, the design of the  \pkg{DALEX}  assumes that each explanation can be presented in Rashomon's perspective. This means that any number of explainers can be placed on a single graph, e.g., for cross-comparison of different models. This helps to see their advantages and disadvantages. 

The \pkg{flashlight} package, similarly to \pkg{DALEX}, allows to combine explanations for various models and plot them together. It also provides a surrogate tree models' explanation. However, the biggest advantage of the \pkg{flashlight} package, not implemented in any other XAI framework covered in this paper, is the possibility to put weights on observations while computing the explanations.  Thus, it is great for applications where observations have sampling weights.

The most distinctive attribute of the \pkg{iml} package is implementation on the basis of R6 classes. Additionally, due to the differences in the method of calculation of SHAP values, it computes that explanation faster than \pkg{DALEX} and \pkg{flashlight}. The \pkg{iml} just like \pkg{flashlight} provides surrogate tree model explanation and model level interactions based on Friedman H-statistics.

\section{Discussion}

The article compared 27 different R packages with methods for XAI analysis of predictive models. The selection criteria (Section~\ref{sec:criteria}) limited the pool of packages that were analyzed. However, we believe that the constraints resulted in the fact that all considered libraries were mature and have a~group of everyday users.  Moreover, the fact that the package is published on CRAN proves that it is operational, has been tested, and is being maintained. 

The first observation refers to explanation methods. Analysis of explanation tasks that are covered by various R packages shows that the overall distribution of those tasks over libraries is changing in time. Back in the day, explanations of predictive models were focused on a global level, with popular profiling methods, such as Partial Dependence, such as (\pkg{pdp}, September 2016) or Accumulated Local Effect plots (\pkg{ALEplot}, November 2017). The Model Parts task also fit that trend with the \pkg{vip} package (June 2018) and variable importance related function in \pkg{pdp}. Nowadays, the Predict Parts Task seems to be more and more popular, especially methods related to Shapley values. On top of previously published packages, such as \pkg{shapper} (March 2019) and \pkg{fastshap} (November 2019), there are new, recently created tools. \CRANpkg{shapr} \citep{shapr},  \CRANpkg{SHAPforxgboost} \citep{SHAPforxgboost} and \pkg{treeshap}, which still awaits publication on CRAN, are examples of such new packages. Not to mention other Predict Parts explanation methods like BreakDown. Another new trend in eXplainable Artificial Intelligence is fairness. There was only one package dedicated to fairness analysis in our comparison, but a number of recently published packages still await their reception. There are, for instance, \CRANpkg{fairmodels}, \CRANpkg{fairml} \citep{fairml} and \CRANpkg{aif360} \citep{aif360} packages. Their number, the short time from publication, and the fact that \pkg{fairness} was first published in September 2019, support the statement that fairness is a new, popular trend in R XAI that cannot be overseen.

The packages that were compared were designed to serve as eXplainable Artificial Intelligence support for model creators and users. However, different tools can be used in the various stages of the model development process. It is easy to explain with the help of an example from our list. The main target group of \pkg{flashlight} users is different than the target of \pkg{modelStudio} or \pkg{modelDown}. The first one is a tool that serves mostly model developers, so they can fit models based on XAI experience, which is an important part of the model life cycle. The second package is dedicated to end-users; those who get a ready model and want to explore its behavior, rather than fitting a new one. That second group of packages is an important new trend in the XAI toolkits world. They are a gateway to model exploration for those who lack expert knowledge about XAI and modeling itself but want to familiarize themselves with the behavior of the model they are using. This trend is also followed by recently developed packages such as \pkg{arenar} or \pkg{xai2cloud} \citep{xai2cloud}.

One more aspect that distinguishes the compared packages is their complexity. On the one hand, some packages focus on one method like \pkg{pdp} or \pkg{ALEPlot}, while, on the other hand, there are packages, such as \pkg{iml} or \pkg{DALEX}, that provide a wide range of different explanation method. There are also tools that present a unique approach to certain areas of XAI. We find it necessary to acknowledge those.
\begin{itemize}
    \item \pkg{DALEXtra} - is designed to bridge XAI packages with frameworks for ML model developments including those coming from different programming languages like Python \pkg{scikit-learn}. It also helps in finding a solution to the Rashomon effect problem, by providing an interface to compare model explanations and performance.
    \item \pkg{flashlight} - this is one of the tools that provides a wide range of explanation methods for complex model analysis. What distinguishes it from similar \pkg{DALEX} and \pkg{iml} packages is weight use cases.~\pkg{flashlight} allows users to specify the weight to every observation and consider them while computing explanations.
    \item \pkg{modelStudio} and \pkg{arenar} - both of these packages provide an easy to use interface to model explanations. Such analysis requires no expert knowledge, since, with a few lines of code, a stand-alone HTML document with precalculated explanations (\pkg{modelStudio}), or an online browser application (\pkg{arenar}) can be created. Moreover, \pkg{arenar} is capable of presenting explanations for more than one model at once and comparing them (See Rashomon Effect in Section~\ref{sec:rashomon}).
    \item \pkg{treeshap} - this is an example of a new, yet-to-be published package that could be an important tool in the future. It uses the additive nature of Shapley values and allows them to be directly computed for ensemble tree-like models. In addition, the implementation in C++ makes the computations fast.
    \item \pkg{triplot} implements explanations that take into account the correlation of variables. This tool, instead of considering a single variable, explains a model using aspects. An aspect is a group of variables (usually correlated ones) that is considered as one variable.
    \item \pkg{vip} - this is a package oriented around variable importance. It proposes an interesting concept of mixing up model-specific and model-agnostic explanation types. For some types of models, a natural variable importance measure can be extracted, while, at the same time, there is an option to access that measure in a model-agnostic way. It provides three different ways of computing model-agnostics variable importance, including permutational variable importance, Shapley-based variable importance, and variance-based variable importance.
    
\end{itemize}

\section{Summary}

In this article, we presented R packages dedicated to eXplainable Artificial Intelligence.  We started demonstrating the importance of XAI in today's world (see Section~\ref{sec:importance}), and we discussed the issue of previously presented taxonomies. We proposed a taxonomy dedicated to predictive model explanations (see Section~\ref{sec:taxonomy}).

Research of the literature has shown that there are plenty of R packages dedicated to XAI. Each of them is extensive to a different degree (Section~\ref{sec:packages_comaprison}), providing a wide range of various opportunities for explaining models using a number of tools. On the one hand, there are packages such as \CRANpkg{ALEplot} \citep{ALEPlot} and \CRANpkg{lime} \citep{limeR} that implement one specific method, or are oriented around one explanation task. On the other hand, packages such as \pkg{flashlight} or \pkg{DALEX} implement diverse methods belonging to different explanation tasks. Explanation of predictive models can also be utilized in the process of finding a model that is better for a given use-case (Section~\ref{sec:rashomon}). This process is simplified by the interfaces provided by some of the frameworks. 
Section~\ref{sec:interoperability} shows that R packages for XAI are flexible in terms of interoperability with frameworks used for model development. This means that XAI methods can be used for any type of predictive model. 

We believe that examples presented in Section \ref{sec:gallery}  and in the \url{xai-tools.drwhy.ai} webpage will encourage more frequent use of XAI tools during the modeling process. The large range of diversity among R packages that provide explanations of machine learning models means that everyone should be able to find a method convenient to use. 

\section{Acknowledgements}

We would like to thank the whole $\text{MI}^{2}\text{DataLab}$ team, especially Hubert Baniecki and Anna Kozak for the discussions and support. Last, but not least, we would like to thank Patrick Hall for his \mbox{valuable~comments}. Work on this paper was financially supported by the NCN Opus grant 2017/27/B/ST6/0130.

%Its purpose is to help everyone concerned understand models' predictions. Good examples of software realizing XAI concept in practice are \CRANpkg{iml} \citep{iml} or \CRANpkg{DALEX} \citep{DALEX} R packages and \pkg{lime} \citep{lime} or \pkg{shap} \citep{NIPS2017_7062} Python libraries. 

\bibliography{Maksymiuk-Gosiewska-Biecek}

\address{Szymon Maksymiuk\\
  Faculty of Mathematics and Information Science\\
  Warsaw University of Technology\\
  Poland\\
  ORCiD: 0000-0002-3120-1601\\
  \email{sz.maksymiuk@gmail.com}}
  
\address{Alicja Gosiewska\\
  Faculty of Mathematics and Information Science\\
  Warsaw University of Technology\\
  Poland\\
  ORCiD: 0000-0001-6563-5742\\
  \email{gosiewska@gmail.com}}

\address{Przemysław Biecek\\
  Faculty of Mathematics and Information Science\\
  Warsaw University of Technology\\
  Faculty of Mathematics, Informatics, and Mechanics\\
  University of Warsaw\\
  Poland\\
%  Samsung Research and Development Institute Poland\\
%  AI Team\\
  ORCiD: 0000-0001-8423-1823\\
  \email{przemyslaw.biecek@gmail.com}}

\clearpage

\appendix

\section{Appendix: Example gallery for XAI packages}
\label{sec:gallery}

The following paragraphs briefly discuss consecutive XAI packages. For each of the packages listed in Table \ref{tab:pkgs_list}, we prepared an example use-case in the form of a \pkg{markdown} document. The documents have similar chapters and sections to facilitate a comparison of the capabilities of each package.
Also, in each use-case, we use the titanic dataset from \pkg{DALEX} package.  It consists of both numerical and categorical dependent variables, and, therefore, it was possible to inspect how toolkits handle different types of features.

The Python library \pkg{aix360} \citep{aix} is a framework that aims at interpretability and explainability of datasets and machine learning models with the help of a method designed for that purpose. It implements, among others, local post-hoc methods like SHAP and LIME, global post-hoc \textit{profWeight}. \pkg{aix360} also provides direct local and global-local explanations (for instance for Generalized Linear Models) as well as methods for explaining the data.

The R package \CRANpkg{ALEPlot} \citep{ALEPlot} creates ALE profiles that demonstrate the behavior of the predictive model response based on one or two dependent variables. The method is described as an improvement to Partial Dependence Profiles and the library can work with any type of model. 
Find an example at \url{http://xai-tools.drwhy.ai/ALEplot.html}.

The R package \CRANpkg{arules} \citep{arules} is dedicated for transaction data analysis. They provide a wide range of methods for rule modeling. \CRANpkg{arulesCBA} \citep{arulescba} for \pkg{arules} is an extension package and provides support for classification glass-box predictive models. Another extension, \CRANpkg{arulesViz} \citep{arulesviz}, equips users with plenty of methods to visualize and inspect a fitted model.
Find an example at \url{http://xai-tools.drwhy.ai/arules.html}.

The R package \CRANpkg{auditor} \citep{auditor} is dedicated for complex model diagnostics. It provides plots and metrics that allow the user to evaluate the performance of a model. On top of that, packages make it possible to conduct complex residual-based model diagnostics. 
Find an example at \url{http://xai-tools.drwhy.ai/auditor.html}.

\begin{figure}[htb]
  \centering
  \includegraphics[scale=0.3]{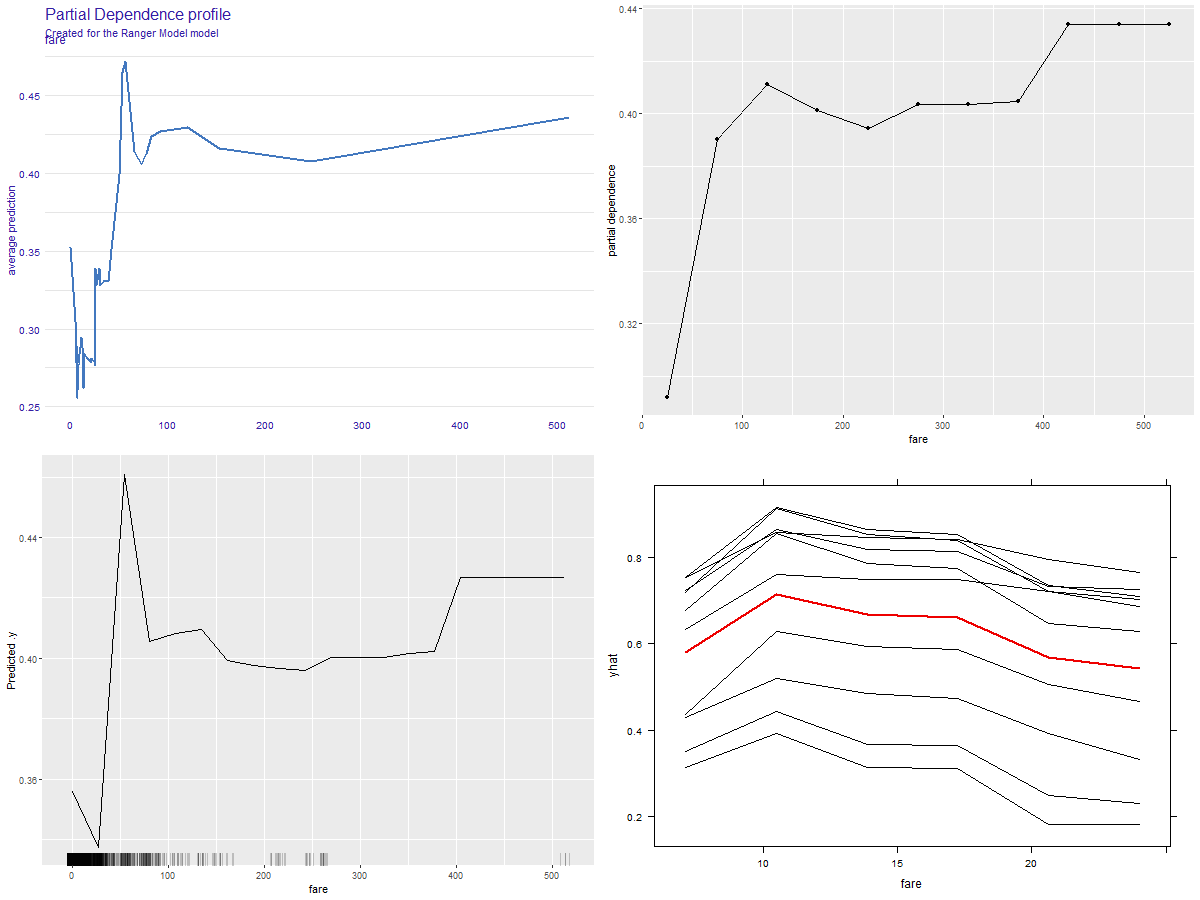}
  \caption{Partial Dependence Porfiles for the fare variable from the \textit{titanic} dataset generated with \pkg{DALEX} (top-left), \pkg{flashlight} (right-top), \pkg{iml} (left-bottom), \pkg{pdp} (right-bottom). We can see that the profiles differs, which is due to the fact that profiles are calculated by default on different grids of points. The computation times of the plots are 0.42s for \pkg{DALEX}, 0.34s for \pkg{flashlight}, 1.88s for \pkg{iml}, 0.03s for \pkg{pdp}.}
  \label{fig:pdp}
\end{figure}

\begin{figure}[htb]
  \centering
  \includegraphics[scale=0.38]{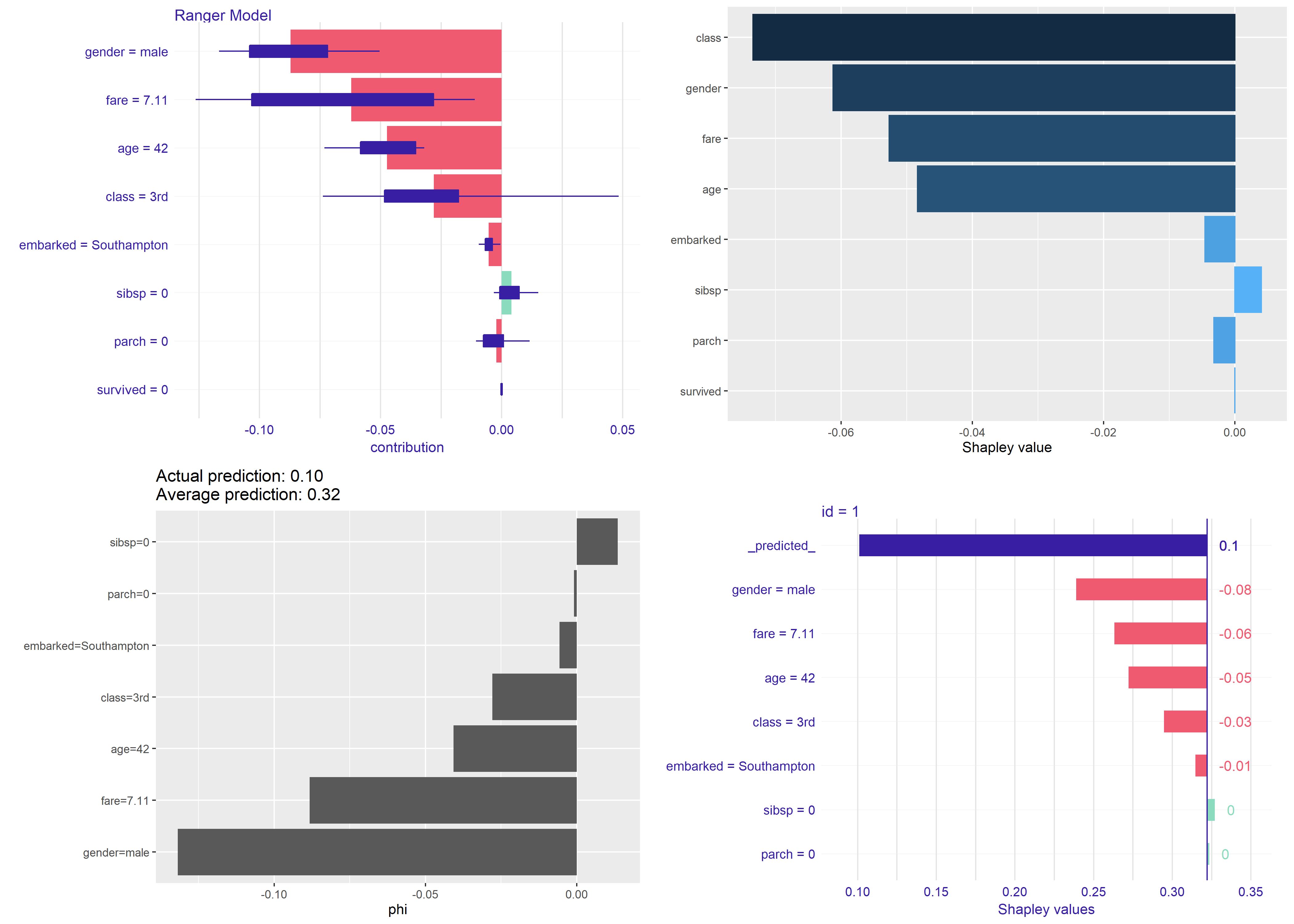}
  \caption{Contribution plots based on Shapley values for the same observation from \textit{titanic} dataset generated with \pkg{DALEX} (top-left), \pkg{fastshap} (top-right), \pkg{iml} (bottom-left), \pkg{shapper} (bottom-right). The computation times of the plots are 41.08s for \pkg{DALEX}, 107.77s for \pkg{fastshap}, 0.39s for \pkg{iml}, 0.02s for \pkg{shapper}.}
  \label{fig:shap}
\end{figure}

The R package \CRANpkg{DALEX} \citep{DALEX} provides  tools for Explanatory Model Analysis \citep{ema}. It represents the global explanation approach by serving variable importance interface, ability to create Partial, Accumulated, and Conditional Dependence Profiles. It facilitates the performance of global and local residual-based model diagnostics. On the local explanations side, it implements SHAP, BreakDown, oscillation contributions, and Ceteris Paribus (ICE). Examples of PDP and SHAP explanations can be seen in the Figure~\ref{fig:pdp} and Figure~\ref{fig:shap}.
Find an example at \url{http://xai-tools.drwhy.ai/DALEX.html}.

The R package \CRANpkg{DALEXtra} \citep{DALEXtra} serves as an extension for \pkg{DALEX}. Its main purpose is to provide a set of pre-defined predict functions for different ML frameworks. This facilitates the integration of XAI packages with modest popular model classes.
Find an example at \url{http://xai-tools.drwhy.ai/DALEXtra.html}.

The R package \CRANpkg{EIX} \citep{EIX} is a model-specific tool designed to explain \pkg{xgboost} models. It provides the ability to diagnose the model by plotting its structure, find variables with the biggest interaction, and perform variable importance using them. It is also possible to explain a single observation using this package.
Find an example at \url{http://xai-tools.drwhy.ai/EIX.html}.

The Python library \pkg{eli5} \citep{eli5} provides two ways to inspect black-boxes: permutation importance and text explanations with LIME. The library \pkg{eli5} supports the most common Python frameworks and packages: scikit-learn, Keras, xgboost, LightGBM, CatBoost, lightning, and sklearn.

The R package \CRANpkg{ExplainPrediction} \citep{ExplainPrediction} allows users to explain predictive model instances through instances showing how values of variables contributed to the prediction of each observation in test data. Explanations can be then aggregated into one global variable importance. 
Find an example at \url{http://xai-tools.drwhy.ai/ExplainPrediction.html}.

The R package \CRANpkg{fairness} \citep{fairness} is dedicated for fairness analysis. It provides 9 different metrics that allow the user to recognize if predicted values contain bias. Plots of metrics and density of probabilities in subgroups are also included.
Find an example at \url{http://xai-tools.drwhy.ai/fairness.html}.

The R package \CRANpkg{fastshap} \citep{fastshap} approximates Shapley values for any type of predictive model. Through this tool variable importance, profiles of variables, and contributions for a single observation can be acquired. Provides an interface for Python \pkg{shap} plots. An example of SHAP can be seen in the Figure~\ref{fig:shap}.
Find an example at \url{http://xai-tools.drwhy.ai/fastshap.html}.

The R package \CRANpkg{flashlight} \citep{flashlight} provides methods that can be used for wide-model analysis, including variable importance, and methods of profiling variables like PDP, ALE, residual, target, and predicted value profiles. A single prediction can be explained with this tool as well with the help of SHAP, BreakDown, and ICE. An example of PDP explanation can be seen in Figure~\ref{fig:pdp}.
Find an example at \url{http://xai-tools.drwhy.ai/flashlight.html}.

The R package \CRANpkg{forestmodel} \citep{forestmodel} generates forest plots of the estimated coefficients of models. The library supports objects produced by \code{lm()}, \code{glm()}, and \code{survival::coxph()} functions. 
Find an example at \url{http://xai-tools.drwhy.ai/forestmodel.html}.

The R package \CRANpkg{fscaret} \citep{fscaret} is associated with \CRANpkg{caret}. It can train various types of predictive models while acquiring variable importance during that process. Any type of model supported by \pkg{caret} can be explained.
Find an example at \url{http://xai-tools.drwhy.ai/fscaret.html}.

The R package \CRANpkg{glmnet} \citep{glmnet} is an interface for fitting Generalized Linear Models. They are examples of glass-box interpretable-by-design models that can be explored via analysis of coefficients provided by the package.  
Find an example at \url{http://xai-tools.drwhy.ai/glmnet.html}.

The R package \CRANpkg{naivebayes} \citep{naivebayes} is a tool dedicated to fitting Naive Bayes predictive models. It provides various types of support for different distributions. It also implements plots that allow users to inspect the model itself. 
Find an example at \url{http://xai-tools.drwhy.ai/naivebayes.html}.

The R package \CRANpkg{iBreakDown} \citep{iBreakDown} provides methods for local model explanations. It allows us to compute and visualize the additive and interaction BreakDown of a~single observation. It also provides an interface for SHAP.
Find an example at \url{http://xai-tools.drwhy.ai/iBreakDown.html}.

The R package \CRANpkg{ICEbox} \citep{ICE} allows user to create ICE curves across the dataset. It also aggregates them creating Partial Dependence Curves or Partial Derivative Curves, and provides an interface for clustering ICE curves. 
Find an example at \url{http://xai-tools.drwhy.ai/ICEbox.html}.

The R package \CRANpkg{iml} \citep{molnar2019} implements a wide range of global and local model-agnostic explanation methods, such as feature importance, PDP, ALE, ICE, surrogate models, LIME, and SHAP. The \pkg{iml} library is characterized by the use of R6 classes. Examples of PDP and SHAP explanations can be seen in Figure~\ref{fig:pdp} and Figure~\ref{fig:shap}. 
Find an example at \url{http://xai-tools.drwhy.ai/iml.html}.

The R package \CRANpkg{ingredients} \citep{ingredients} implements techniques for both local and global explanations. It facilitates the computation of permutational variable importance, Accumulated, Partial, and Conditional dependence profiles. The package provides a tool for computing Ceteris Paribus (ICE) curves, clustering them, and calculating oscillations in order to explain a single variable.
Find an example at \url{http://xai-tools.drwhy.ai/ingredients.html}.

\begin{figure}[h!]
  \centering
  \includegraphics[scale=0.4]{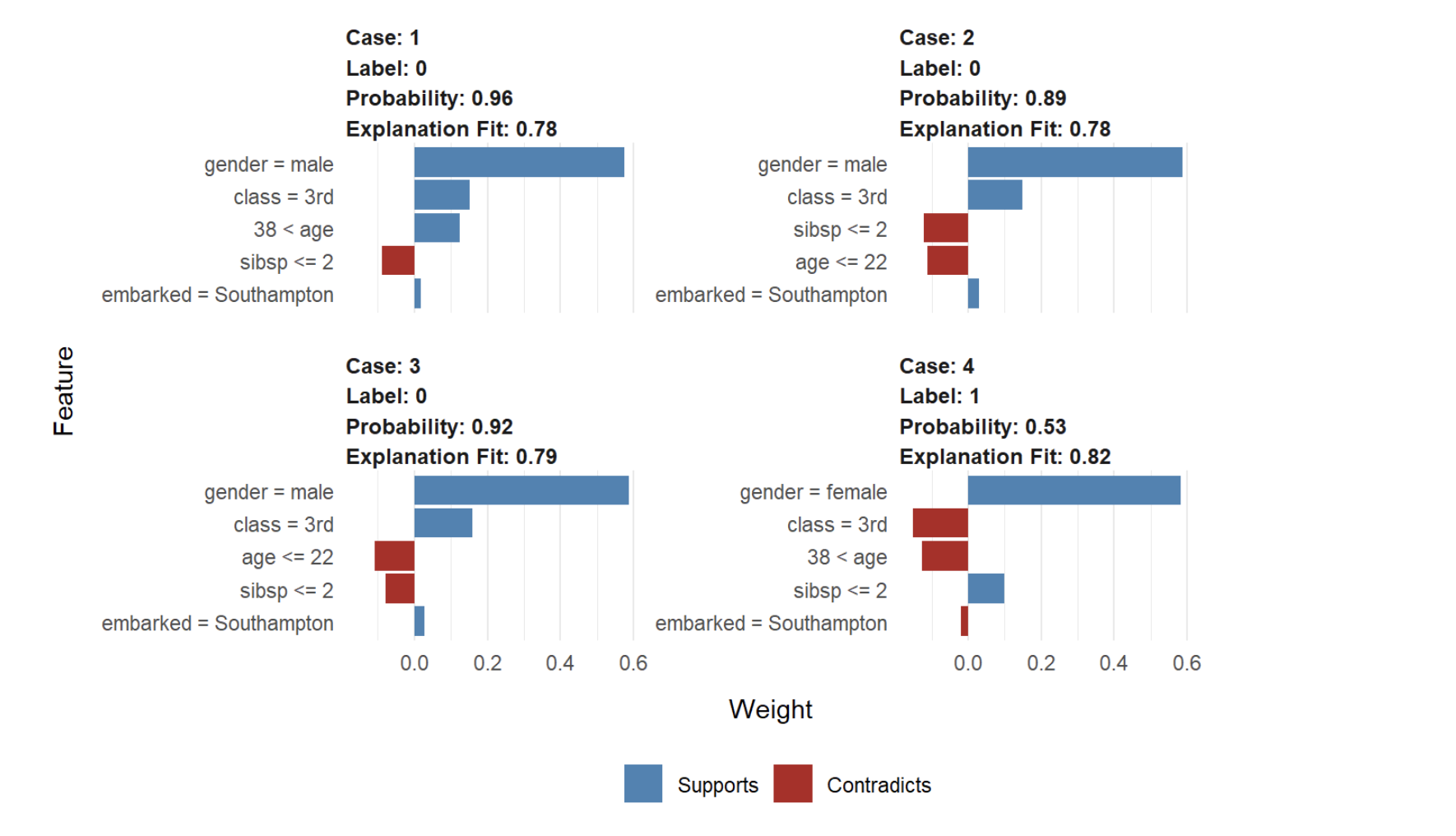}
  \caption{Snapshot of the explanations created with the \pkg{lime} package.}
  \label{fig:lime}
\end{figure}

An R package \pkg{interpret} \citep{interpretml} is a tool for training Explainable Boosting Machine models \citep{10.1145/2783258.2788613} that are high-performance generalized additive models with pairwise interactions. The Python version (library \pkg{interpret}) implements more interpretable models, which are decision trees, linear regression, and logistic regression. It also provides XAI methods, such as SHAP, Tree SHAP, LIME, Morris Sensitivity Analysis, and PDP. 

The R package \CRANpkg{kknn} \citep{kknn} provides an extended interface for training classifiers based on the k-Nearest Neighbors method. The package provides ways to inspect the nearest neighbors (the most similar observations).
Find an example at \url{https://mi2datalab.github.io/XAIL-tools/kknn.html}.

The Python library \pkg{lime} is an implementation of the LIME \citep{lime} technique by the authors of this method. The library supports text, image, and tabular data explanations. 

The R package \CRANpkg{lime} \citep{limeR} is an implementation that is independent of the authors of the original Python library. An R tool supports a wide range of frameworks, e.g. caret, parsnip, and mlr. See an example in Figure \ref{fig:lime}. 
Find an example at \url{http://xai-tools.drwhy.ai/lime.html}.

The R package \CRANpkg{live}  \citep{RJ-2018-072} provides local, interpretable, and model-agnostic visual explanations. The idea behind LIVE is similar to LIME, the difference is the definition of the surroundings of an observation. A neighborhood of the observation of interest is simulated by perturbing one variable at a time. Therefore, numerical variables are used in the interpretable local model and are not discretized as in the basic LIME. 
Find an example at \url{http://xai-tools.drwhy.ai/live.html}.

The R package \CRANpkg{mcr} \citep{mcr} is a tool to compare two measurement methods using regression analysis. The package contains functions for summarizing and plotting results. The \textit{mcr} provide comparisons of models, yet is limited only to the regression models.
Find an example at \url{http://xai-tools.drwhy.ai/mcr.html}.

The R package \CRANpkg{modelDown} \citep{modelDown} generates a website with HTML summaries for predictive models.  The generated website provides information about model performance, variable response (PDP, ALE), and the importance of variables, and concept drift. The \pkg{modelDown} also provides a comparison of models. Additionally, data available on the website can be downloaded and recreated in the R session.
Find an example at \url{http://xai-tools.drwhy.ai/modelDown.html}.

\begin{figure}[htb]
  \centering
  \includegraphics[scale=0.4]{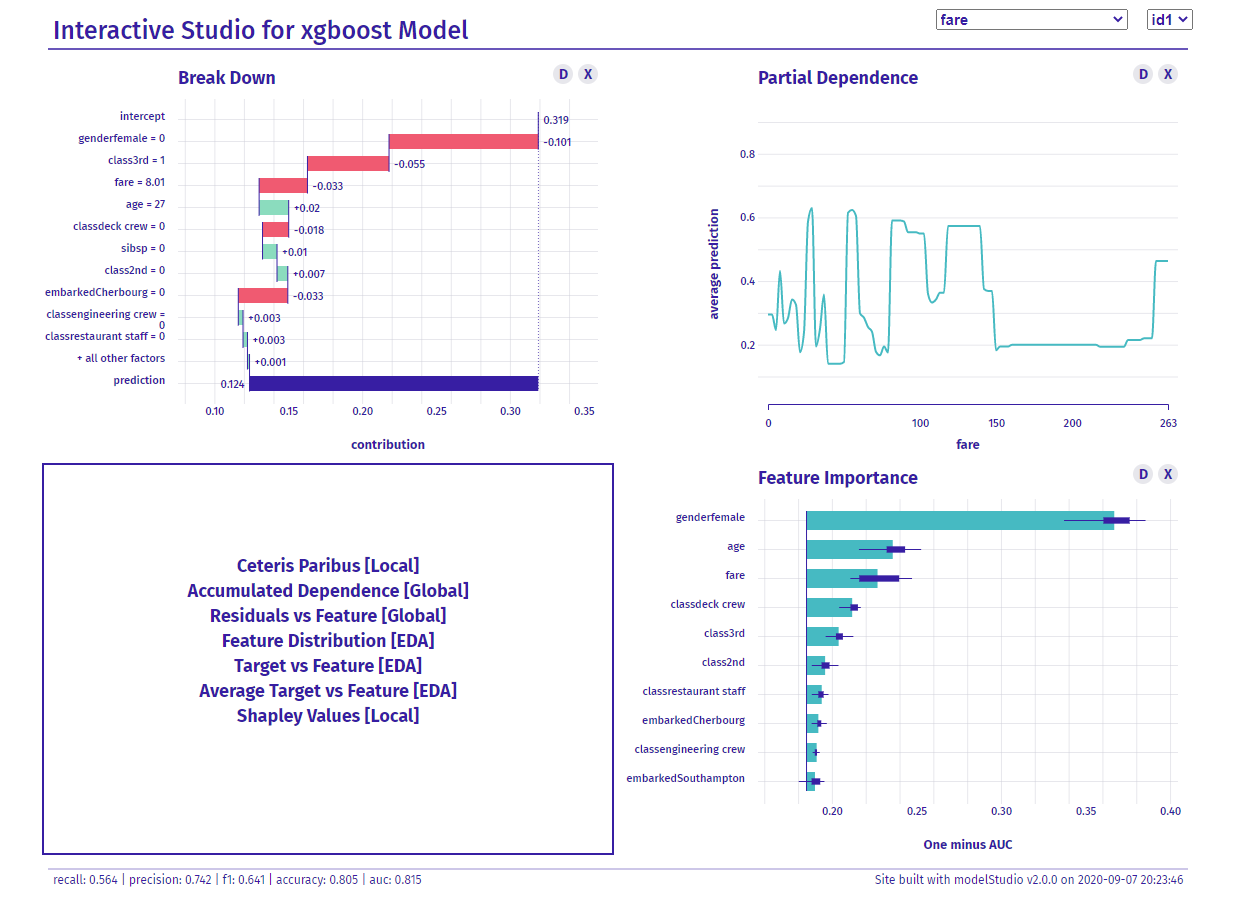}
  \caption{Snapshot of the HTML site created by \pkg{modelStudio}.}
  \label{fig:ms}
\end{figure}

The R package \CRANpkg{modelStudio} \citep{modelStudio} generates interactive and animated model explanations in the form of a serverless HTML site. modelStudio provides various explanations, such as SHAP, Ceteris Paribus, permutational variable importance, PDP, ALE, and plots for exploratory data analysis. The package can be easily integrated with the \textit{scikit-learn} and \textit{lightgbm} Python libraries. An example screenshot of the HTML site with explanations is presented in Figure~\ref{fig:ms}.
Find an example at \url{http://xai-tools.drwhy.ai/modelStudio.html}.

The R package \CRANpkg{party} \citep{partyctree} is a tool dedicated for recursive partitioning. The core of the package is the implementation of the decision tree glass-box model. Decision paths can be plotted using it, along with the split rules. The visualization also covers the distribution of values in terminal nodes.
Find an example at \url{http://xai-tools.drwhy.ai/party.html}.

The R package \CRANpkg{pdp} \citep{pdp} is a tool for computing PDP and ICE. This library works with any predictive model. An example of the PDP explanation can be seen in Figure~\ref{fig:pdp}.
Find an example at \url{http://xai-tools.drwhy.ai/pdp.html}.

The R package \CRANpkg{randomForestExplainer} \citep{randomForestExplainer} is a set of tools for explaining random forests. The existing explanations show variable importance, distribution of minimal depth for each variable, variable interactions, and prediction plot for two variables. What is more, a package provides an option to generate all explanations as an HTML report. 
Find an example at \url{http://xai-tools.drwhy.ai/randomForestExplainer.html}.

The Python library \pkg{shap} is an implementation of the Shapley-based explanations technique (SHAP) \citep{NIPS2017_7062} provided by its authors. SHAP explanations are supported by many visualizations. The library provides also a high-speed algorithm for tree-based models and SHAP-based variable importance and detection of variable interactions.

The R package \CRANpkg{shapper} \citep{shapper} is an interface for Python library \pkg{shap}. A~package implements new plots, different than in the Python version. The \pkg{shapper} provides plotting explanations for multiple models together. An example of the SHAP explanation can be seen in Figure~\ref{fig:shap}.
Find an example at \url{http://xai-tools.drwhy.ai/shapper.html}.

 The Python library \pkg{Skater} \citep{skater} is a tool for global and explanations. The implemented features are: variable importance, partial dependence profiles, LIME, Scalable Bayesian Rule Lists, Tree Surrogates, and two methods for deep neural networks, i.e. Layer-wise Relevance Propagation (e-LRP), and Integrated Gradient.

The R package \CRANpkg{smbinning} \citep{smbinning} provides tools to organize the end-to-end development process of building a scoring model. The library covers data exploration, variable selection, feature engineering, binning, and model selection. 
Find an example at \url{http://xai-tools.drwhy.ai/smbinning.html}.

The R package \CRANpkg{survxai} \citep{survxai} is, to the best of our knowledge, the only tool for model-agnostic explanations of survival analysis models. The package implements local and global explanations, and it also provides comparisons of models.
Find an example at \url{http://xai-tools.drwhy.ai/survxai.html}.

The R package \CRANpkg{vip} \citep{vip} provides many ways to calculate variable importance and interaction strength measures. There is model-based variable importance for models such as random forest, gradient boosted decision trees and multivariate adaptive regression splines. The \pkg{vip} package also provides three model-agnostic variable importance measures: permutation-based, Shapley-based, and variance-based. 
Find an example at \url{http://xai-tools.drwhy.ai/vip.html}.

The R package \CRANpkg{vivo} \citep{vivo} is an implementation of variable importance measures. Global importance is based on oscillations of partial dependence profiles while local importance is based on oscillations of Ceteris Paribus profiles. 
Find an example at \url{http://xai-tools.drwhy.ai/vivo.html}.

\end{article}

\end{document}